\documentclass[journal]{IEEEtran}

\newif\ifanonymous
\anonymousfalse

\usepackage{amsmath} 
\usepackage{graphicx}
\usepackage{booktabs}
\usepackage{makecell}
\usepackage[utf8]{inputenc}
\usepackage[T1]{fontenc}
\usepackage{url}
\usepackage{balance}
\usepackage{tabularx}
\usepackage{subfigure}

\title{
SwarmCoDe: A Scalable Co-Design Framework for Heterogeneous Robot Swarms via Dynamic Speciation
}

\ifanonymous
\author{Anonymous Authors}
\else
\author{Andrew Wilhelm$^{1*}$ and Josie Hughes$^{1}$%
\thanks{$^{*}$Correspondence to
{\tt\small andrew.wilhelm@epfl.ch}}
\thanks{$^{1}$Department of Mechanical Engineering, École Polytechnique Fédérale de Lausanne}%
}
\fi

\ifdefined
\ifanonymous
\else
\fi
\else
\ifdefined\XeTeXversion
\ifanonymous
\AtBeginDocument{}
\else
\AtBeginDocument{}
\fi
\fi
\fi

\begin{document}
\bstctlcite{IEEEexample:BSTcontrol} 

\maketitle

\begin{abstract}
Robot swarm co-design requires optimizing not only individual robot designs and behaviors, but also the heterogeneous swarm structure: how many robot types should exist, how they specialize and cooperate, and in what proportions they are deployed. However, these structural choices are often specified in advance rather than optimized.
We present SwarmCoDe, a Cooperative Co-Evolutionary Algorithm (CCEA) that incorporates swarm decomposition into the optimization process without requiring the number of species to be specified in advance.
SwarmCoDe combines dynamic speciation with evolved genetic tags, partner selectivity, and relative dominance to jointly optimize robot hardware, task-planning behavior, cooperative relationships, and deployed swarm composition.
Across replicated simulated foraging experiments, SwarmCoDe discovers task-dependent swarm decompositions and adapts them when task requirements change, in contrast to fixed-$K$ CCEAs which are sensitive to their pre-defined decomposition.
We further show that fabrication budgets can change the preferred swarm composition and robot designs, and that SwarmCoDe can adapt its co-design under these resource constraints.
Finally, SwarmCoDe co-designs 200-agent swarms using only 50 evolutionary individuals, demonstrating deployment at four times the evolutionary population size.
Together, these results demonstrate that the heterogeneous structure of a robot swarm can be incorporated as a design variable and optimized alongside the robots themselves.

\end{abstract}


\vspace{-1mm}
\section{INTRODUCTION}

Swarm robotics is an increasingly important area of autonomous systems, enabling capabilities that can exceed those of single- or multi-agent platforms~\cite{brambillaSwarmRoboticsReview2013}. 
By leveraging emergent behavior, robotic swarms can collaboratively accomplish large-scale, complex tasks that would otherwise be intractable.
Furthermore, swarms are inherently robust; losing a member of the swarm does not drastically affect the performance of the entire swarm.

\begin{figure}[!thbp]
    \centering
    \includegraphics[scale=0.3]{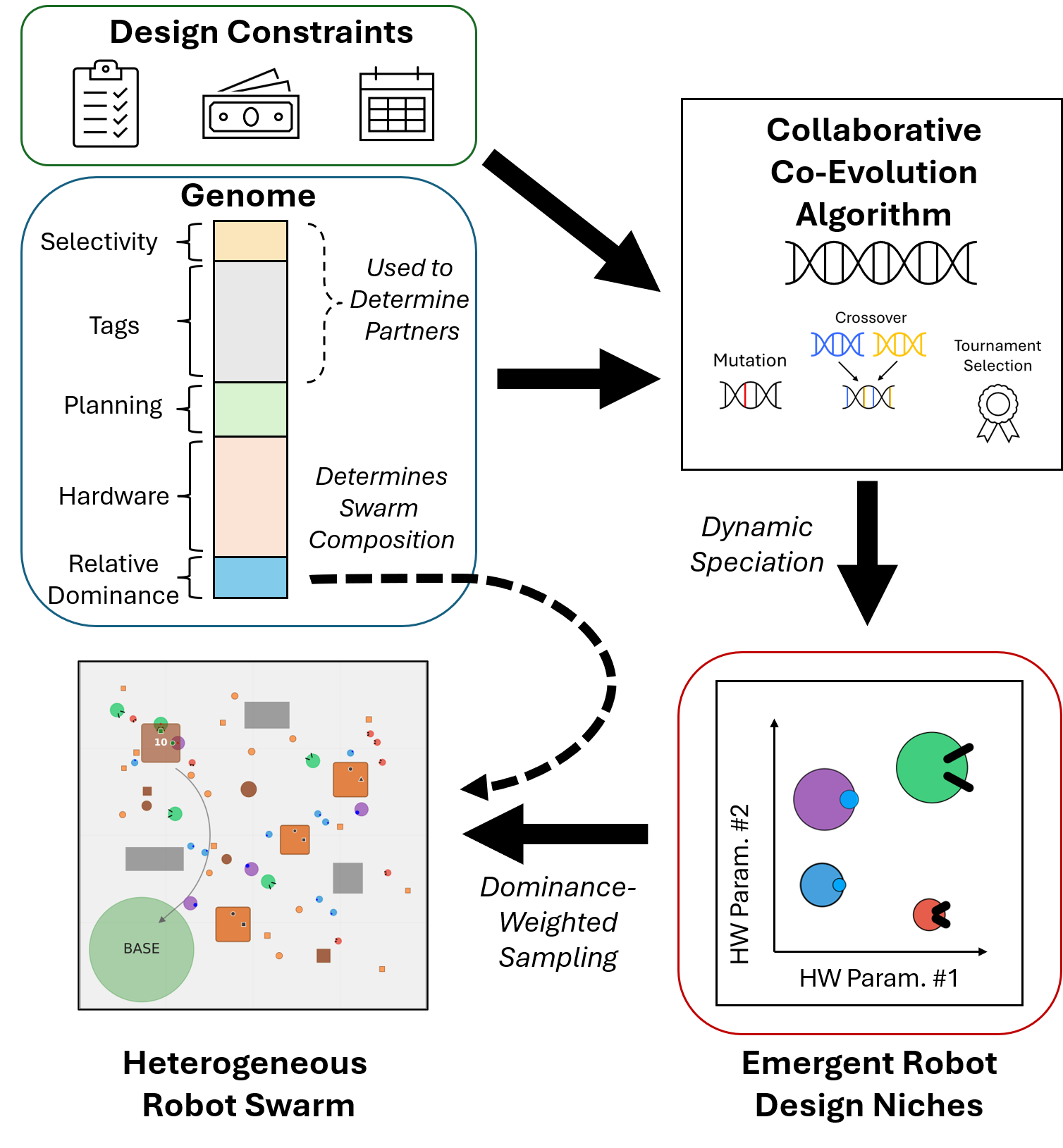}
    \vspace{-5mm}
    \caption[Caption for LOF]{
    Overview of the SwarmCoDe framework for heterogeneous robot swarm co-design.
    SwarmCoDe evolves task-planning and hardware-design variables alongside genetic tags, partner selectivity, and relative dominance.
    Dynamic speciation allows distinct robot species to emerge during optimization, while the tags and partner-selectivity gene govern which species cooperate.
    Relative dominance determines the representation of selected robot designs in the deployed swarm, decoupling swarm size from the evolutionary population.
    Together, these mechanisms allow the robot designs and heterogeneous structure of the swarm to be determined jointly through co-design.
    }
    \label{fig:overview_figure}
\end{figure}

Consequently, the optimization of swarm robotic systems via co-design (i.e., the concurrent design across multiple domains such as morphology, control, component selection, task planning, etc.) holds significant potential, although it has been a largely under-researched area~\cite{liCoevolutionFrameworkSwarm2015}. 
At the scale of hundreds or thousands of agents, even minor improvements to individual robot performance compound into massive gains in collective efficiency. 
In addition, when fabricating an extensive number of robots, minimizing per-unit cost and build time becomes paramount. 
While a marginal reduction in fabrication time or component cost may seem inconsequential for a single robot, it is critically important for the economic feasibility of large-scale swarm deployments.

However, the same scale that makes co-design especially important for robot swarms also makes these systems difficult to design.
Beyond the traditional challenges of robot design (multi-domain constraints, mixed continuous and discrete variables, and multi-objective optimization~\cite{johanssonComponentBasedOptimization2007, spielbergFunctionalCooptimizationArticulated2017}), swarm co-design has its own additional challenges. 
The design space is extremely non-intuitive, as it is not clear how certain changes in one robot will propagate through the swarm and affect emergent collaboration. 
Moreover, the size of the design space itself scales exponentially with the size of the swarm.
These additional challenges to swarm co-design motivate the need for swarm-specific co-design algorithms as opposed to naïvely applying single or multi-robot design algorithms to these systems.

For heterogeneous swarms, there is an additional structural challenge: the optimizer must determine not only the physical and behavioral design of the robots, but also how the swarm itself should be decomposed into distinct robot types. 
The number of useful robot types, their roles, their cooperative relationships, and their relative prevalence may all depend on the task and engineering constraints. 
Cooperative Co-Evolutionary Algorithms (CCEAs) provide a natural approach for optimizing heterogeneous teams, but many formulations use a fixed problem decomposition or predefined subpopulations~\cite{maSurveyCooperativeCoEvolutionary2019}.
This can be restrictive when the appropriate number of robot types is not known \textit{a priori} or changes with the task and engineering constraints.
Adaptive decomposition has also been explored in cooperative co-evolution, including species creation and removal based on evolutionary stagnation and component contribution~\cite{potterCooperativeCoevolutionArchitecture2000} and dynamic merging and splitting of robot-controller populations~\cite{gomesDynamicTeamHeterogeneity2018}.
For heterogeneous swarm co-design, however, the decomposition is also coupled to the robots' physical designs, behavioral policies, cooperative relationships, and deployed composition.

In this paper, we propose SwarmCoDe, a novel, species-based Cooperative Co-Evolutionary Algorithm (CCEA) explicitly formulated for robot swarm co-design (Figure~\ref{fig:overview_figure}). 
Our approach emergently finds unique robot ``species'', distinct robot designs that specialize in different collaborative roles within the swarm. 
Rather than prescribing the number of species in advance, dynamic speciation allows the decomposition of the swarm to emerge during optimization. 
Evolved genetic tags, inspired by biological signaling cues such as floral coloration that attracts pollinators, and a partner-selectivity gene determine cooperative relationships among species, while a relative dominance gene determines their relative representation in the deployed swarm.
By stochastically sampling swarm members in proportion to their evolved dominance genes, our algorithm can scale to swarms that are larger than the number of individuals in our evolutionary pool. 
Together, these mechanisms allow SwarmCoDe to jointly optimize robot hardware, task-planning behavior, cooperative organization, and deployed swarm composition.

We demonstrate that our SwarmCoDe algorithm can co-design robot swarms composed of up to 200 agents for simulated collaborative foraging tasks, optimizing behavior and hardware (morphology and component selection). 
We first show that the degree of heterogeneity discovered by SwarmCoDe changes with the functional structure of the task.
Comparisons with fixed-$K$ CCEAs show that performance is sensitive to the decomposition specified \textit{a priori} and that the best-performing $K$ varies across tasks, whereas SwarmCoDe requires no predefined $K$.
We further demonstrate that SwarmCoDe adapts its species structure when task requirements change during evolution and use ablation studies to characterize the mechanisms underlying specialization, cooperation, and swarm composition. 
Finally, fabrication-budget experiments show that the useful degree of specialization changes with available resources, while large-swarm experiments demonstrate co-design of deployed swarms containing up to 200 agents using an evolutionary population substantially smaller than the deployed swarm.
Together, these results demonstrate that heterogeneous swarm structure can itself be treated as an engineering co-design variable, allowing the number, characteristics, and composition of robot types to adapt alongside their physical and behavioral designs.

\section{RELATED WORK}

\subsection{Robot Co-Design}

Developing optimal robotic systems requires a co-design approach that optimizes performance objectives while minimizing resource consumption under multi-domain constraints, a task complicated by subtle interdependencies across various engineering domains.
Existing literature addresses single-agent co-design through approaches including heuristics~\cite{haComputationalDesignRobotic2018}, grammars~\cite{zhaoRoboGrammarGraphGrammar2020}, gradient-based methods~\cite{spielbergLearningInTheLoopOptimizationEndToEnd2019}, mixed-integer linear programming~\cite{magnussenMulticopterDesignOptimization2015,milojevicCODEIResourceEfficientTaskDriven2025}, constraint programming~\cite{wilhelmConstraintProgrammingComponentLevel2023}, and evolutionary morphology-control co-optimization~\cite{lipsonAutomaticDesignManufacture2000,cheneyScalableCooptimizationMorphology2018}.

However, research into the co-design of multi-agent and swarm systems remains limited. 
Analytical methods such as integer programming~\cite{carloneRobotCodesignMonotone2019}, monotone co-design problems~\cite{zardiniCoDesignAVEnabledMobility2020}, or constraint programming~\cite{wilhelmMonotoneSubsystemDecomposition2025} provide provable guarantees on the optimality of solutions. 
These frameworks have designed multi-agent transport teams, vehicle transportation fleets, and quadcopter fleets, respectively. 
However, to maintain computational tractability, these methods frequently rely on modeling approximations such as linearization or monotonization. 
Because they do not utilize simulation, these formula-based approaches struggle to capture the complex interactions characteristic of swarm systems such as collision dynamics and emergent collaborative behaviors.

\subsection{Genetic Algorithms for Swarm Systems}

Automatic swarm design encompasses modular, evolutionary, and learning-based approaches to control-software design~\cite{francescaAutoMoDeNovelApproach2014,birattariAutomaticOffLineDesign2019,kucklingRecentTrendsRobot2023}, with some approaches also jointly configuring hardware and swarm size under economic constraints~\cite{salmanConcurrentDesignControl2019}.
Evolutionary methods typically evaluate candidate swarm designs in simulation and use their fitness to guide optimization.
Most existing literature on genetic algorithms for swarm systems focuses on optimizing controllers for either homogeneous swarms (all agents identical, using the same controller)~\cite{gomesEvolutionSwarmRobotics2013, ferranteEvolutionSelfOrganizedTask2015} or heterogeneous swarms (agents utilize one or several controllers)~\cite{dambrosioGenerativeEncodingMultiagent2008,eibenCollectiveSpecializationEvolutionary2007, vandiggelenEmergenceSpecialisedCollective2024}, or compares the two~\cite{waibelGeneticTeamComposition2009}.
Physical heterogeneity has also been explored in swarm systems composed of predefined robot types~\cite{dorigoSwarmanoidNovelConcept2013}.

The primary advantage of simulation-based genetic algorithms is their ability to capture complex interactions and emergent behaviors between agents and their environment. 
However, this capability comes at the cost of computational tractability: evaluating the fitness function is typically the most computationally demanding phase of the algorithm. 
Consequently, modern frameworks utilizing hardware acceleration and parallelization are increasingly required to address this bottleneck. 
Furthermore, because these algorithms lack mathematical guarantees, they may converge on locally optimal solutions instead of globally optimal ones.

\subsection{Cooperative Co-Evolution and Dynamic Speciation}

To optimize multi-agent systems where distinct specializations are desired, Cooperative Co-Evolutionary Algorithms (CCEAs) are frequently employed. 
CCEAs evolve multiple subpopulations, where different subpopulations represent different variables, decomposing an $N$-dimensional problem into $N$ cooperating subproblems~\cite{potterEvolvingNeuralNetworks1995}.
Many CCEA formulations use a decomposition determined \textit{a priori}, with separate subpopulations assigned to predefined agents or robot types~\cite{yongCoevolutionRoleBasedCooperation2009}.
In multi-agent applications, this requires the designer to specify aspects of the team decomposition before optimization and can couple the evolutionary representation to the prescribed team structure.
Such formulations also commonly maintain genetic separation between subpopulations, helping preserve distinct specializations.


Other CCEAs address these limitations by determining the co-evolutionary structure automatically, such as creating and removing species during optimization based on evolutionary stagnation and contribution~\cite{potterCooperativeCoevolutionArchitecture2000} or identifying interactions among variables to construct the decomposition~\cite{maSurveyCooperativeCoEvolutionary2019,omidvarCooperativeCoEvolutionDifferential2014}.
Prior work has used cooperative co-evolution to encourage different robot controllers to specialize in complementary roles within a team~\cite{nitschkeEvolvingTeamBehaviors2012}.
Other approaches extend this idea by adapting the degree of team heterogeneity itself, dynamically merging populations according to behavioral characterizations and stochastically splitting them to explore additional specializations~\cite{gomesDynamicTeamHeterogeneity2018}.  
These methods demonstrate that both specialization and team structure can adapt during cooperative co-evolution, but changes in the decomposition are mediated by explicitly defined population structures, behavioral characterizations, or restructuring operations.  
In contrast, SwarmCoDe does not explicitly merge or split populations or require a separate behavioral characterization for restructuring; species emerge directly through compatibility-based speciation as genomes evolve, while evolved genetic tags and partner selectivity separately determine which species cooperate during evaluation.

Speciation and niche preservation also play a broader role in maintaining diversity during evolutionary search.
Species-based evolutionary algorithms maintain multiple optima in multimodal problems~\cite{liSpeciesBasedEvolutionary2010}, while quality-diversity methods maintain diverse, high-performing solutions across behavioral niches~\cite{pughQualityDiversityNew2016}.
NEAT uses speciation to protect new network topologies from premature elimination~\cite{stanleyEvolvingNeuralNetworks2002}.
In these approaches, species or niches represent different regions of the search or behavior space, rather than different cooperating optimization subproblems.

In this paper, we adapt NEAT's compatibility-based speciation to a CCEA framework, where species act as cooperating subpopulations rather than competing explorations of the search space, allowing stable cooperative niches to form without \textit{a priori} species boundaries.
This allows the number of distinct robot types and their respective morphologies and task-planning policies to emerge jointly during evolution, without explicit population merge and split operations based on a separately defined behavioral characterization.
Furthermore, we decouple the evolutionary population structure from the deployed swarm composition by introducing a dominance gene (a continuous value between 0 and 1) within the genome of each individual that determines its relative frequency within the swarm. 
The final swarm is then populated through random sampling weighted by these dominance values.

\section{METHODS}
We consider heterogeneous swarm co-design problems in which multiple robot designs must cooperate to complete shared tasks while satisfying physical and resource constraints. 
The optimization must therefore determine not only the hardware and task-planning behavior of the robots, but also the heterogeneous structure of the deployed swarm. 
In this work, we evaluate this problem using a collaborative foraging task, which provides a controlled setting for varying functional roles, collaborative requirements, and hardware constraints.

\subsection{SwarmCoDe Framework}
\label{sec:framework}
SwarmCoDe is an adaptive cooperative co-evolutionary framework that jointly evolves candidate solutions, their cooperative relationships, an emergent population decomposition, and their relative representation in constructed teams.
This architecture employs a multi-level selection pressure: at the inter-species level, species compete for population slots (offspring allocation) based on their total adjusted fitness, while at the intra-species level, individuals compete for reproduction via tournament selection.
At the same time, individuals must collaborate with members of other species during task evaluation, creating coupled competitive and cooperative evolutionary pressures.

SwarmCoDe separates three aspects of this cooperative organization. 
Dynamic speciation determines the evolutionary population structure; evolved partner selectivity determines which species collaborate during a particular evaluation; and relative dominance determines the prevalence of selected designs in the deployed team. 
Together, these mechanisms allow both the evolutionary decomposition and the composition of cooperative teams to emerge during optimization rather than requiring a fixed number of species or predefined team structure.

A single generation proceeds as follows. 
First, the population undergoes dynamic speciation, where species are assigned and established. 
Next, each individual is evaluated by identifying cooperative partner species, constructing a deployed team, and estimating the individual's marginal contribution through paired focal and baseline evaluations. 
Finally, species-level performance determines offspring allocation, while parent selection and evolutionary operators generate the subsequent population. 
We elaborate on each of these steps in the following sections.

\subsubsection{Genome Structure}

\begin{table}[t]
\centering
\caption{SwarmCoDe genome components and their roles.}
\label{tab:genome_components}
\footnotesize
\setlength{\tabcolsep}{4pt}
\renewcommand{\arraystretch}{1.05}
\begin{tabularx}{\columnwidth}{@{}
    >{\raggedright\arraybackslash}p{0.36\columnwidth}
    >{\raggedright\arraybackslash}X
@{}}
\toprule
\multicolumn{1}{c}{\textbf{Genome Component}} &
\multicolumn{1}{c}{\textbf{Role in SwarmCoDe}} \\
\midrule

Robot co-design variables $g^{\mathrm{design}}$
& Physical and behavioral candidate robot design \\

Genetic tag $g^{\mathrm{tag}}$
& Evolutionary compatibility and cooperative signaling \\

Partner selectivity $g^{\mathrm{selectivity}}$
& Determines acceptable cooperative partner species \\

Relative dominance $g^{\mathrm{dominance}}$
& Determines relative representation in the deployed swarm \\

\bottomrule
\end{tabularx}
\end{table}

Each SwarmCoDe individual contains two classes of variables: robot co-design variables, which define the candidate robot's physical and behavioral design, and cooperative-organization variables, which govern how candidate designs are organized and deployed within the heterogeneous swarm. 
The specific robot co-design variables may vary with the swarm-design problem, whereas SwarmCoDe introduces a genetic tag, partner-selectivity gene, and relative-dominance gene to govern speciation, cooperative partner selection, and deployed swarm composition.
The genome can therefore be represented as
$ g = \left[ g^{\mathrm{design}}, g^{\mathrm{tag}}, g^{\mathrm{selectivity}}, g^{\mathrm{dominance}} \right]$.
The role of each genome component is summarized in Table~\ref{tab:genome_components}.

The genetic tag is a binary vector that contributes both to evolutionary compatibility and to cooperative partner selection. 
The partner-selectivity gene determines how similar another species' tag must be for that species to be accepted as a collaborator. 
The relative-dominance gene determines the relative representation of selected robot designs when constructing the deployed swarm.

Importantly, neither species identity nor the number of species \(K\) is directly encoded in the genome. 
Instead, the evolutionary species structure emerges through the dynamic-speciation procedure described below.

For the foraging-based robot swarm co-design problems studied in this paper, \(g^{\mathrm{design}}\) contains task-planning and hardware/morphology variables. 
These application-specific variables are detailed in Section~\ref{sec:heterogeneous_codesign}.

\subsubsection{Dynamic Speciation}

Dynamic speciation allows the evolutionary decomposition of the swarm to emerge and change during optimization.
First, individuals in the population must be assigned to species.
This is done in a manner similar to the species assignment in NEAT.
A prototype representing the species is identified as the individual closest to the previous generation's prototype.
Then, the remaining individuals are iterated through and compared against the existing species prototypes.
If an individual's genetic distance to a prototype is below the speciation threshold $\delta$, it is assigned to the first compatible species encountered.
If no existing prototype lies within $\delta$, the individual establishes a new species.
Because species membership is reassigned each generation, individuals that diverge beyond the threshold can establish new species, while previously distinct species can be grouped back into a common species as their members become sufficiently similar.
Species can also disappear as evolutionary selection reduces poorly performing species until they have no remaining members, allowing the number of active species to increase or decrease throughout optimization.

SwarmCoDe requires a compatibility measure $D$ that captures meaningful differences among the candidate solutions being co-designed.
During speciation, this distance is compared against the threshold $\delta$ to determine whether individuals belong to the same species.
The specific compatibility metric can therefore be adapted to the co-design representation.
For the robot hardware and task-planning representation used in this paper, we define $D$ using differences in the genetic tag, hardware, task-planning policy, end effector, and robot size, as detailed in Section~\ref{sec:heterogeneous_codesign}.

\subsubsection{Evaluation and Fitness}

\begin{figure*}[t]
\vspace{2mm}
\centering
\includegraphics[scale=0.42]{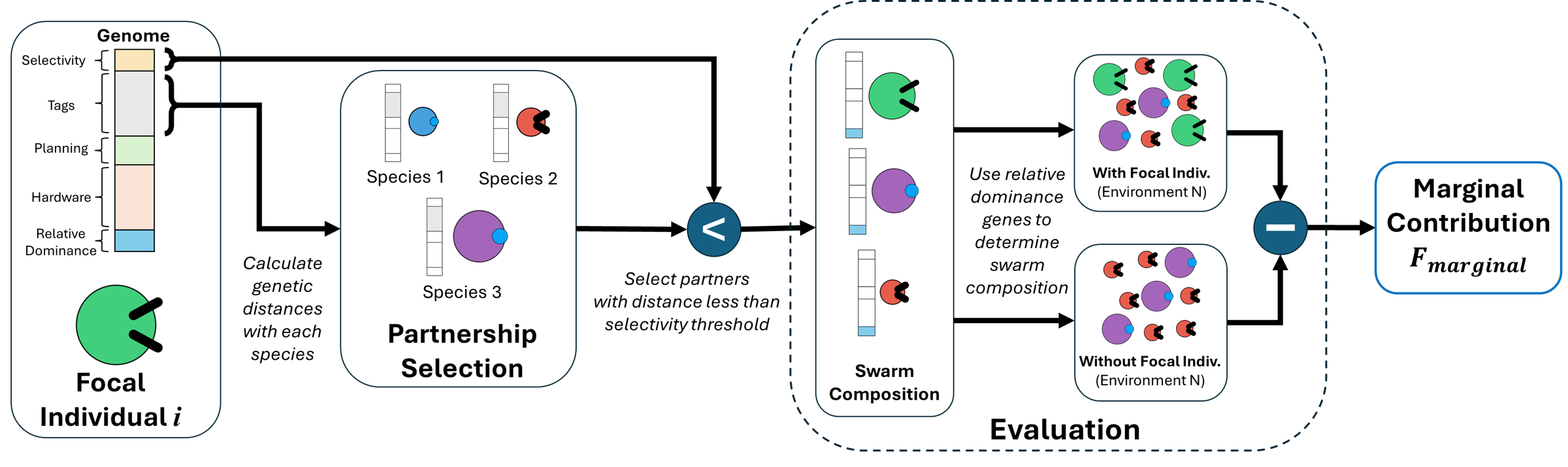}
\vspace{-4mm}
\caption[Caption for LOF]{
Evaluation pipeline for calculating marginal contribution. 
Swarms are stochastically assembled using the evolved dominance gene to determine swarm composition ratios. 
By comparing the performance of a swarm containing the focal individual against a baseline swarm where that individual is replaced by selected partner representatives in an identical environment, the paired evaluation estimates the focal individual's marginal contribution under the defined team-construction and environment distribution.
}
\label{fig:evaluation_figure}
\vspace{-5mm}
\end{figure*}

During evaluation, every individual in the population is assessed in turn, see Figure~\ref{fig:evaluation_figure}.

\textbf{Partner Selection and Swarm Composition: }
An individual selects its partners using its genetic tag, a set of binary values of length $L$ in its genome.
The distance between the focal individual's tags and the prototype tag of each other species is computed.
Each individual possesses an evolved \textit{partner-selectivity gene} with a value between 0 and 1, which defines the maximum allowable tag distance for a species to be considered a partner.
If the distance between tags is less than this selectivity threshold, the species becomes a partner.
Decoupling partner collaboration from overall genetic distance allows species that are genetically similar to still fulfill different niches on a team.
This tag-and-threshold rule is structurally similar to tag-based cooperation models, in which an individual's tolerance determines how similar another agent's tag must be for cooperation~\cite{rioloEvolutionCooperationReciprocity2001}.
Here, the rule operates at the level of species-partner eligibility rather than directly determining an agent's cooperative behavior.
Finally, for each species identified as a partner, one representative is sampled uniformly from a pool of up to five members of that species for each evaluation episode. 
This pool includes the individuals retained as elites from the previous generation, with any remaining positions filled by randomly selecting other members of the current species.

The focal individual and selected partner representatives define the candidate robot designs from which the deployed swarm is assembled.
First, the focal individual and each partner species are guaranteed at least one representative in the swarm.
The remaining swarm slots are filled using clones of the focal individual or the selected partner representatives, utilizing their respective \textit{relative dominance genes} as relative selection probabilities.
We refer to this evolved variable as relative dominance because it controls relative representation among the selected robot designs rather than specifying an independent absolute probability.
Dominance is an evolved continuous value between 0 and 1 dictating the individual's desired representation ratio within the swarm.
This dominance gene allows the algorithm to scale to scenarios where the number of individuals in the evolutionary population is less than the number of agents in the physical swarm, particularly useful for swarm co-design where agents number in the hundreds.

\textbf{Marginal-Contribution Fitness: }
To assign cooperative credit to each candidate robot design, we conduct two sets of evaluation trials: one with the focal individual and one without.
This follows the same general principle as difference evaluation functions, which estimate an individual's contribution by comparing overall team performance with a corresponding baseline in which that individual's contribution is removed or replaced~\cite{wolpertOptimalPayoffFunctions2001,agoginoEfficientEvaluationFunctions2008}.
This paired evaluation is intended to isolate the contribution of the focal design while reducing variance caused by unrelated environmental stochasticity.

For the focal evaluation, the focal individual and its selected partners are evaluated across randomly generated environmental realizations, with a new deployed swarm sampled from the relative dominance gene values for each realization.
Averaging over these trials reduces fitness noise from both stochastic initial conditions and stochastic swarm composition, so that the resulting fitness reflects the selected designs and their relative dominance rather than any single sampled swarm.

A second set of baseline trials is conducted under identical environmental conditions but with the focal individual removed.
To maintain a consistent swarm size, the slots previously occupied by the focal individual are repopulated through random sampling of the same partner representatives, again using their dominance genes as relative probabilities.


The marginal contribution $F_{\mathrm{marginal}}$ is calculated by taking the difference between these sets of evaluations $F_{\mathrm{marginal}} = F_{\mathrm{focal}} - F_{\mathrm{baseline}}$. 
This quantity estimates the focal individual's marginal contribution under the defined collaborator and environment distribution rather than its ``true fitness''.

We use a gated fitness to then assign the focal individual its fitness score.
Rather than using the magnitude of the marginal contribution directly as the fitness, as is typical in difference evaluations, we use it as a gate on team fitness.
If the marginal contribution is positive, then the fitness score remains unchanged.
Otherwise, the fitness score is penalized by multiplying by a scalar $P_{\mathrm{marginal}}<1$.
This preserves selection pressure toward overall swarm performance while penalizing individuals that do not provide a positive contribution to the team, and avoids making fitness overly sensitive to marginal magnitude, which can vary with the number of deployed swarm slots occupied by a design.

\subsubsection{Evolutionary Operators}

These operators are used to determine the population and genetic material for the subsequent generation.

\textbf{Elitism and Reproductive Selection: }We utilize per-species elitism to preserve the individuals with the highest fitness scores in each species across generations. 
The number of elites retained by species $s$ is $N^{\mathrm{elite}}_s=\min\left(E,\lfloor n_s/5\rfloor+1\right)$
where $n_s$ is the number of individuals currently assigned to the species and $E = 5$ is the maximum permitted elite count per species.
Because $\lfloor n_s/5 \rfloor + 1 \geq 1$, every species retains at least its best member, while the cap $E$ limits elite carryover.
This prevents large species from completely dominating the population and preserves slots for subsequent exploration.

\textbf{Offspring Allocation and Parent Selection: }Species are assigned an offspring quota proportional to their total adjusted fitness, rewarding successful species with higher representation.
Available population slots are allocated proportionately to these adjusted fitness values.

To fulfill these quotas, parents are selected via tournament selection.
Generally, mating is restricted to individuals within the same species.
The first parent is selected through a tournament within the target species.
A second tournament then identifies a potential second parent; however, this individual is only accepted if the genetic distance between the two parents remains below the threshold $\delta$.
If this threshold is violated, the second tournament is repeated for up to 5 attempts.
In cases where a compatible intra-species second parent cannot be found after these attempts, a global tournament is conducted to facilitate inter-species mating.

Following selection, mating occurs.
If parents belong to the same species, there is a 70\% probability of crossover and a 30\% probability of cloning the first parent.
If species differ, the crossover probability is reduced to 2.5\%.

\textbf{Mutations:} Every generated offspring undergoes mutation to maintain genetic diversity and facilitate exploration of the design space.
Exactly one tag bit, chosen uniformly at random, is flipped, and the partner-selectivity and relative-dominance genes are perturbed by Gaussian noise with standard deviation $\sigma = 0.1$ and clipped to $[0,1]$.
Mutation operators for the application-specific task-planning and hardware variables are detailed in Sections~\ref{sec:control} and~\ref{sec:hardware}, respectively.


\subsection{Heterogeneous Robot Swarm Co-Design}
\label{sec:heterogeneous_codesign}

\begin{table}[ht]
\centering
\footnotesize
\setlength{\tabcolsep}{4pt}
\caption{Genome structure and parameter bounds}
\label{tab:genome_structure}
\begin{tabularx}{\columnwidth}{@{} l l l X @{}}
\toprule
\textbf{Component} & \textbf{Type} & \textbf{Domain} & \textbf{Phenotypic Expression} \\
\midrule
BT Opcodes       & Int[]   & $[0, 13]$               & Behavior tree encoding \\
Radius           & Float   & $[0.3, 2.0]\,\text{m}$  & Size and carrying capacity \\
Motor Torque     & Float   & $[0.8, 6.0]\,\text{N·m}$  & Acceleration and lifting limit \\ 
Battery Capacity & Float   & $[50, 200]\,\text{J}$   & Total stored energy \\
End Effector     & Categ.  & $\{0, 1\}$              & Suction vs. pincher hardware \\
Motor Tier       & Categ.  & $\{0, 1, 2\}$           & Torque-to-mass vs. cost \\
Battery Tier     & Categ.  & $\{0, 1, 2\}$           & Energy-to-mass vs. cost \\
Chassis Tier     & Categ.  & $\{0, 1, 2\}$           & Chassis mass density vs. cost \\
\bottomrule
\end{tabularx}
\end{table}

We deploy SwarmCoDe for the co-design of robot swarms, concurrently optimizing both task-planning policies and hardware morphology.
In the experiments considered here, each evolutionary individual therefore represents a reusable robot design together with its task-planning policy. 
The robot co-design variables \(g^{\mathrm{design}}\) introduced in Section~\ref{sec:framework} encode these physical and behavioral properties, while the tag, partner-selectivity, and relative-dominance variables determine how candidate designs are organized, cooperate, and are represented within the deployed swarm.
The application-specific robot co-design variables and their parameter bounds are summarized in Table~\ref{tab:genome_structure}.

The swarm is tasked with foraging for packages of varying weights and sizes dispersed throughout the environment.
Successful retrieval requires robots with compatible physical capabilities and end effectors, while collaborative packages additionally require multiple robots to coordinate for collective transport.

\subsubsection{Simulation Environment}
 We utilize a custom-developed, 2D physics engine optimized for GPU acceleration using vectorized parallel updates in JAX.
The engine employs semi-implicit Euler integration and resolves elastic collisions between robots and static obstacles.
Robots interact with individual and collaborative packages whose tool, size, weight, and location requirements are varied across the experimental conditions described in Section~\ref{sec:implementation_details}.
We represent each candidate robot through two application-specific components: its task-planning policy and its hardware design, described in the following section.

\subsubsection{Planning}
\label{sec:control}
We utilize a behavior tree to encode the task planning policy.
Within the genome, this is encoded as an array of instructions, where each instruction consists of an opcode and a jump index.
These opcodes are strictly categorized into three functional types: control flow, environmental conditions (sensing), and physical actions.
The instruction set is detailed in Table~\ref{tab:bt_opcodes}.

Standard behavior trees rely on recursive node traversal, which is highly inefficient for vectorized GPU execution.
To support efficient vectorized evaluation, we encode each tree as a flattened, fixed-length array where control flow is managed through opcodes and jump indices. 
Shorter behavior trees are padded using \texttt{NOP} opcodes.
At the beginning of each simulation tick, an interpreter traverses this array sequentially.
Action and condition nodes return standard state flags (SUCCESS, FAILURE, or RUNNING).
Physical locomotion actions, such as navigating toward a target, are executed via an underlying PD controller.
Control nodes (\texttt{SEQ}, \texttt{SEL}) evaluate these flags; if a \texttt{SEQ} child fails or a \texttt{SEL} child succeeds, the interpreter utilizes the pre-computed jump index to immediately bypass the remaining subtree.
The \texttt{END} opcode explicitly signals the termination boundary of these subtrees.

To bootstrap the optimization, the initial behavior tree for each individual is drawn at random from a small library of 6 hand-designed foraging templates.
Mutations are 95\% point mutations and 5\% subtree replacements, and jump indices are recomputed following any genetic alteration.

\begin{table}[ht]
\centering
\caption{Opcodes used for the Behavior Tree representation}
\label{tab:bt_opcodes}
\resizebox{\columnwidth}{!}{%
\begin{tabular}[t]{rl}
\toprule
\textbf{Opcode} & \textbf{Name} \\
\midrule
0 & \texttt{SEQ} (Sequence) \\
1 & \texttt{SEL} (Selector) \\
\midrule
2 & \texttt{COND\_HAS\_PACKAGE} \\
3 & \texttt{COND\_NEAR\_PACKAGE} \\
4 & \texttt{COND\_NEAR\_BASE} \\
5 & \texttt{COND\_AM\_I\_STUCK} \\
\bottomrule
\end{tabular}
\hspace{1em}
\begin{tabular}[t]{rl}
\toprule
\textbf{Opcode} & \textbf{Name} \\
\midrule
6 & \texttt{ACT\_MOVE\_TO\_PACK} \\
7 & \texttt{ACT\_MOVE\_TO\_BASE} \\
8 & \texttt{ACT\_RANDOM\_WALK} \\
9 & \texttt{ACT\_PICK\_UP} \\
10 & \texttt{ACT\_DROP} \\
11 & \texttt{ACT\_MOVE\_TO\_RANDOM\_PACK} \\
\midrule
12 & \texttt{END} \\
13 & \texttt{NOP} \\
\bottomrule
\end{tabular}
}

\end{table}

\subsubsection{Hardware Design}
\label{sec:hardware}
The hardware design space encompasses both continuous morphological parameters and discrete categorical components.
Specifically, the evolutionary process optimizes the continuous radius of the robot, which governs its spatial footprint and carrying capacity, alongside motor torque and battery capacity operating points.
Categorical genes define the functional hardware configuration, including the equipped end effector type and discrete performance tiers for the chassis material, battery, and motor.
This structure creates functional trade-offs where premium tiers provide superior structural integrity or energy density but incur higher costs.
These choices consequently affect package eligibility, carrying capability, energy use, morphology, and fabrication cost, coupling individual robot design to the performance and economics of the overall swarm.
By evolving these continuous and discrete hardware parameters simultaneously alongside the task-planning policy, SwarmCoDe can discover physically and behaviorally distinct robot designs specialized for complementary roles within the swarm.

During mutation, each continuous hardware gene (radius, motor torque, and battery capacity) is perturbed with probability $p=0.15$ by Gaussian noise with standard deviation equal to 10\% of its parameter range and clipped to the bounds in Table~\ref{tab:genome_structure}.
The binary end-effector gene is flipped with a reduced probability of $0.1p$ to preserve tool specialization.
Each component-tier gene mutates with probability $0.15p$, shifting to an adjacent tier with probability 0.7 or being resampled uniformly from its domain otherwise.

\textbf{Compatibility Metric:}
For dynamic speciation, we define an application-specific compatibility metric over the robot design representation described above.
The genetic compatibility distance $D$ between two genomes is calculated as a weighted sum of distance metrics across the relevant gene components:
\begin{equation}
   \begin{split}
   D ={}& w_{\mathrm{tag}} d_{\mathrm{tag}} + w_{\mathrm{hw}} d_{\mathrm{hw}} + w_{\mathrm{BT}} d_{\mathrm{BT}} \\
   &+ w_{\mathrm{tool}} d_{\mathrm{tool}} + w_{\mathrm{size}} d_{\mathrm{size}}
   \end{split}
\end{equation}

Here, $d_{\mathrm{tag}} = (H/L)^\gamma$, where $H$ is the Hamming distance between tag bitstrings, $L$ is the tag length, and $\gamma$ is a scaling exponent.
The term $d_{\mathrm{hw}}$ represents the normalized Euclidean distance in the continuous hardware space, excluding the radius and tool parameters.
The behavior tree (BT) difference $d_{\mathrm{BT}}$ is the fraction of differing opcodes within the behavior tree array.
Morphological differences are captured by $d_{\mathrm{tool}} \in \{0, 1\}$, which penalizes differing end effectors, and $d_{\mathrm{size}}$, the normalized absolute difference in chassis radius.
The component distance metrics are summarized in Table~\ref{tab:distance_weights}, while their weighting coefficients are reported in Table~\ref{tab:hyperparameters}.

\begin{table}[ht]
\centering
\small
\caption{Application-specific distance metrics used to compute genetic compatibility}
\label{tab:distance_weights}
\begin{tabular}{lll}
\toprule
\textbf{Component} & \textbf{Distance} & \textbf{Metric} \\
\midrule
Tag & $d_{\mathrm{tag}}$ & $(H/L)^\gamma$ \\
Hardware & $d_{\mathrm{hw}}$ & Normalized Euclidean distance \\
Control (BT) & $d_{\mathrm{BT}}$ & Fraction of differing opcodes \\
Tool & $d_{\mathrm{tool}}$ & Binary difference ($0$ or $1$) \\
Size & $d_{\mathrm{size}}$ & Normalized $\Delta$ radius \\
\bottomrule
\end{tabular}
\vspace{-5mm}
\end{table}

\subsubsection{Task Variants}
We evaluate SwarmCoDe across several variants of the foraging task to test how the evolved swarm structure changes with the task requirements.
In the one-niche task, all packages require a pincher end effector.
In the two-niche task, packages require either a pincher or suction end effector, creating two distinct functional roles.
In the four-niche task, both package types are used and package weights decrease with distance from the base.
This creates separate near/heavy and far/light roles for each end effector, resulting in four functional niches.

We additionally modify these task conditions to evaluate specific aspects of the framework.
In the dynamic task, the environment transitions from the two-niche to the four-niche condition during evolution.
An asymmetric two-niche task is used to evaluate whether the relative-dominance gene can adapt the deployed swarm composition when the two roles are not equally valuable.
Finally, fabrication-budget and scalability experiments evaluate how the evolved swarm design changes under resource constraints and as the deployed swarm size increases.

\subsubsection{Fitness Function}

The fitness function is designed to prioritize successful package delivery while providing intermediate rewards for partial task progress and encouraging energy-efficient behavior.
The raw fitness is calculated as:
\begin{equation*}
\begin{split}
F_{\mathrm{raw}} = \max \left( \right. & (S_{\mathrm{delivery}} + S_{\mathrm{collab}} + S_{\mathrm{pickup}} + S_{\mathrm{energy}} \\
& + S_{\mathrm{proximity}} + S_{\mathrm{closeness}}) \times P_{\mathrm{activity}}, \left. 0.1 \right)
\end{split}
\end{equation*}
The individual fitness terms are defined as:
\begin{itemize}
    \item $P_{\mathrm{activity}}$ is an inactivity penalty. If zero packages are retrieved: $P_{\mathrm{activity}} = 0.5$. Otherwise, $P_{\mathrm{activity}} = 1.0$
    \item $S_{\mathrm{delivery}} = N_{\mathrm{delivered}} \times W_{\mathrm{delivery}}$ rewards the successful retrieval of packages back to the base
    \item $S_{\mathrm{collab}} = (\sum G_{\mathrm{delivered}}) \times W_{\mathrm{delivery}} + N_{\mathrm{collab\_delivered}} \times W_{\mathrm{collab\_bonus}}$ applies a bonus for collaborative packages, where  $\sum G_{\mathrm{delivered}}$ is the total required grip points across delivered collaborative packages
    \item $S_{\mathrm{pickup}} = (N_{\mathrm{picked}} + N_{\mathrm{collab\_picked}}) \times W_{\mathrm{pickup}}$ provides a minor reward for initial acquisition to guide early-stage evolution
    \item $S_{\mathrm{energy}} = E_{\mathrm{avg\_final}} \times W_{\mathrm{energy}}$ is a small boost to encourage robots to conserve energy
    \item $S_{\mathrm{proximity}} = \text{mean}(Score_i) \times W_{\mathrm{proximity}}$ guides robots already holding a package towards base, where $Score_i = \frac{10}{1 + 0.1 \times d_{\mathrm{base}}}$ if holding and $0$ otherwise for each robot $i$ in the swarm
    \item $S_{\mathrm{closeness}} = \text{mean}(P_i) \times W_{\mathrm{closeness}}$ rewards intermediate progress for packages moved but not delivered, where $P_{i} = \frac{D_{\mathrm{initial}} - D_{\mathrm{final}}}{D_{\mathrm{initial}}}$ for package $i$
\end{itemize}

Aggregate hardware costs are regulated by a swarm-level budget constraint $C_{\mathrm{budget}}$.
The total swarm cost $C_{\mathrm{swarm}}$ includes the hardware cost of the deployed robots and, for the fabrication-budget experiments, an additional fabrication cost of \$500 for each unique design in the swarm.
We enforce this using a soft penalty $P_{\mathrm{budget}}$:
$$
P_{\mathrm{budget}} = 
\begin{cases} 
\max(0.05, e^{-\lambda \Delta C}) & \text{if } \Delta C > 0 \\ 
1 & \text{otherwise} 
\end{cases}
$$
where $\Delta C = C_{\mathrm{swarm}} - C_{\mathrm{budget}}$ represents the budget excess, and $\lambda$ is a decay coefficient that dictates the strictness of the soft penalty for exceeding $C_{\mathrm{budget}}$.
The soft penalty has a slower decrease (unlike a hard penalty that immediately penalizes above the threshold) that helps guide the evolution towards swarm designs that satisfy the budget constraint.

The final fitness score is $F = F_{\mathrm{raw}} \times P_{\mathrm{budget}}$, smoothed with an EMA filter (decay rate $\alpha$).

We additionally consider a return-on-investment (ROI) objective in which no fabrication budget is imposed. 
Instead, the optimizer directly maximizes performance relative to swarm cost:
$$ F_{\mathrm{ROI}}=\frac{F_{\mathrm{raw}}}{C_{\mathrm{swarm}}}. $$
This differs from the fixed-budget formulation: the budget experiments maximize performance subject to a specified resource budget, whereas the ROI objective allows the economically preferred operating cost to emerge during optimization.
This formulation is useful when a deployment is not constrained by a fixed budget, but the designer still wants to identify whether additional swarm performance justifies the added hardware and fabrication cost.

\subsection{Experimental Protocol and Implementation}
\label{sec:implementation_details}

\subsubsection{Experimental Protocol}

\begin{table}[t]
\centering
\caption{Summary of experiment parameters}
\label{tab:experiment_protocol}
\footnotesize
\setlength{\tabcolsep}{3.5pt}
\begin{tabular}{lccc}
\toprule
\textbf{Experiment} &
$\mathbf{N_{\mathrm{pop}}}$ &
$\mathbf{N_{\mathrm{swarm}}}$ &
\textbf{Generations} \\
\midrule
Static task complexity (1--2 niches) & 100 & 20 & 300 \\
Static task complexity (4 niches)    & 100 & 20 & 500 \\
Fixed-$K$ CCEA         & 100 & 20 & 500 \\
Ablations              & 100 & 20 & 500 \\
Dynamic transition     & 100 & 20 & 750 \\
Asymmetric composition & 100 & 20 & 250 \\
Fabrication budget     & 100 & 20 & 500 \\
Large-swarm scaling    & \makecell{50, 75, 100,\\150, 200} & 200 & 600 \\
\bottomrule
\end{tabular}


\end{table}

Unless otherwise stated, experiments use ten independent evolutionary runs.
Each run uses a different random seed while keeping the algorithmic hyperparameters fixed within a given experimental comparison.
Table~\ref{tab:experiment_protocol} summarizes the evolutionary population size, deployed swarm size, and number of generations used for each experiment.
The static tasks with one or two functional niches are run for 300 generations because these lower-complexity conditions typically stabilize earlier, while the four-niche task is extended to 500 generations to allow additional optimization in the larger search space.

During individual fitness evaluation, each focal individual and its selected partners are evaluated across 25 environmental realizations, with a new swarm composition sampled in each, as described in Section~\ref{sec:framework}.
The large-swarm experiments use 12 environmental realizations per fitness evaluation.
The focal and baseline evaluations use the same environmental realizations, reducing fitness differences caused by unrelated environmental stochasticity.

For comparisons between experimental conditions, we report the best fitness found over each independent evolutionary run.
If $F_{\mathrm{gen}}(g)$ is the fitness of the best team found at generation $g$, the endpoint fitness for a run of $G$ generations is
$$
F_{\mathrm{endpoint}} = \max_{1 \leq g \leq G} F_{\mathrm{gen}}(g).
$$
For convergence plots, we instead report the best fitness found up to each generation,
$$
F_{\mathrm{best}}(g) = \max_{1 \leq k \leq g} F_{\mathrm{gen}}(k).
$$
Because the goal of each evolutionary run is to identify the highest-performing swarm design, we use the running-best fitness to show how the quality of the best solution found improves over the course of evolution.
Similarly, when characterizing swarm structure, we report the number of species represented in the best-performing deployed team rather than the total number of species in the evolutionary population. 
This better reflects the heterogeneous structure actually selected for deployment, since not all species present in the population contribute to the best-performing swarm.

\subsubsection{Baseline and Ablations}

\textbf{Fixed-$K$ CCEA Baseline:} We compare SwarmCoDe with a fixed-$K$ CCEA baseline, where $K$ specifies the number of reproductively isolated evolutionary subpopulations established before optimization.
The total evolutionary population is held at $N_{\mathrm{pop}}=100$ and divided as evenly as possible among the $K$ subpopulations, while the deployed swarm contains $N_{\mathrm{swarm}}=20$ robots.
Subpopulation sizes remain fixed throughout evolution, with selection and crossover performed within each subpopulation.

For each focal evaluation, one elite representative is sampled from every other subpopulation, and the remaining swarm slots are filled by uniform sampling with replacement from the focal individual and these representatives.
We evaluate $K\in\{2,3,4,6,10\}$ on the static two and four-niche tasks using ten independent 500-generation runs per condition.
These experiments use the same robot representation, simulation, task rewards, marginal-contribution fitness, and \$8,000 swarm budget as the matched SwarmCoDe experiments.
The \$500 per-design fabrication cost is removed from both the fixed-$K$ CCEA runs and the matched SwarmCoDe runs, so that larger values of $K$ are not penalized for having more predefined species.

We additionally evaluate $K=20$, where each subpopulation contributes one robot to the 20-robot deployed swarm. 
In this test, removing a single focal robot typically produces only a small change in total swarm performance, making the marginal-contribution gate less informative. 
We therefore use mean swarm fitness for this condition and construct the deployed swarm from the best individual in each subpopulation.

\textbf{Dynamic Task Transition:} To evaluate whether the swarm decomposition can adapt when the task changes during evolution, we run 750-generation experiments that transition from the two-niche task to the four-niche task at generation 250.
We compare SwarmCoDe with fixed-$K$ CCEA for $K\in\{2,3,4,6,10\}$, using ten runs per condition with $N_{\mathrm{pop}}=100$ and $N_{\mathrm{swarm}}=20$.
Evolution continues from the existing population through the transition, and the provided convergence plots compute separate running-best fitness values before and after the transition.
For the dynamic-transition analysis, we quantify changes in the functional decomposition of the running-best deployed team by counting distinct morphological niches.
Each niche is defined by the robot's end effector and size class.
We separate the size classes at a chassis radius of 0.95\,m, the midpoint between the minimum radii required to carry the light (0.5\,m) and heavy (1.4\,m) packages, resulting in at most four possible niches.

\textbf{Ablations:} We evaluate the main components of SwarmCoDe on the four-niche collaborative task using ten independent 500-generation runs per condition.
All ablation conditions and the full SwarmCoDe reference use the \$500 per-design fabrication cost under the \$8,000 budget, matching the fabrication-budget experiments (Section~\ref{sec:budget_results}).
Absolute fitness values in the ablations are therefore lower than in Fig.~\ref{fig:ccea_comparison}(b), but conditions within the ablation study are directly comparable.
Distance Speciation replaces the standard compatibility measure with normalized morphological distance for species assignment and partner selection.
This provides a direct similarity measure, similar to prior adaptive approaches that organize populations using predefined behavioral characterizations~\cite{gomesDynamicTeamHeterogeneity2018}.
Random Partnering retains the standard prototype-based species assignment, but removes selective partner matching during evaluation.
Every active species other than the focal species can contribute a collaborator, sampled uniformly from the same representative pool as SwarmCoDe, and the compatibility restriction on interspecies mating is removed.
No Marginal Fitness removes the focal-versus-baseline contribution evaluation and instead assigns fitness directly from swarm performance.

To more directly test the relative-dominance gene, we use an asymmetric two-niche task in which the two functional roles occur at different frequencies.
In the symmetric task conditions, approximately uniform swarm compositions can perform well, making it difficult to isolate the effect of evolved relative dominance from the other mechanisms determining team performance. 
The asymmetric task contains 60 individual packages, where pincher and suction requirements are sampled with probabilities 0.30 and 0.70, respectively. 
Both package types otherwise have identical mass, size, and spawn distributions, such that the 70/30 split represents the expected task demand rather than an imposed robot composition.
We compare dominance-weighted and uniform swarm sampling using ten independent 250-generation runs per condition and report best-found team fitness, together with the designs and expected composition of each run's best-found team.

\textbf{Large-Swarm Scalability:}
To evaluate whether SwarmCoDe can design deployed swarms that are substantially larger than the evolutionary population, we fix $N_{\mathrm{swarm}}=200$ and vary $N_{\mathrm{pop}}\in\{50,75,100,150,200\}$.
These conditions range from four deployed robots per evolutionary individual to a one-to-one mapping, with evolved designs instantiated multiple times according to their relative-dominance values when $N_{\mathrm{pop}}<N_{\mathrm{swarm}}$.
To accommodate the larger swarm, we use a scaled version of the two-niche task with a larger arena, a fabrication budget of \$80,000, and 112 total packages of which 32 are collaborative packages.
Fitness values are therefore not directly comparable to those of the 20-robot experiments.
All runs proceed for 600 generations, and each focal and baseline fitness evaluation uses 12 environmental realizations.
We use five independent runs for each population size and evaluate best-found team performance together with the composition of the resulting swarms.

\textbf{Statistical Reporting and Implementation:} Across replicated experiments, we show the individual evolutionary seeds where practical together with the mean across runs. 
Error bars and shaded regions show the 95\% confidence interval (CI) for the mean across independent runs.

We detail the hyperparameters used during our experimental evaluation in Table~\ref{tab:hyperparameters}.


\begin{table}[ht]
\centering
\scriptsize
\caption{Hyperparameters and empirical constants used in evaluation}
\label{tab:hyperparameters}
\begin{tabularx}{\columnwidth}{l l l}
\toprule
\textbf{Category} & \textbf{Symbol} & \textbf{Value} \\
\midrule
Distance Weights & $w_{\mathrm{tag}}, w_{\mathrm{hw}}, w_{\mathrm{BT}}$ & $1.0, 0.5, 0.3$ \\
                 & $w_{\mathrm{tool}}, w_{\mathrm{size}}$      & $0.35, 0.7$ \\
\addlinespace
Fitness Weights  & $W_{\mathrm{delivery}}, W_{\mathrm{collab\_bonus}}$ & $100.0, 50.0$ \\
                 & $W_{\mathrm{pickup}}, W_{\mathrm{energy}}$          & $1.0, 0.03$ \\
                 & $W_{\mathrm{proximity}}, W_{\mathrm{closeness}}$    & $1.0, 30.0$ \\
\addlinespace
Constants        & Speciation Threshold $\delta$ & $0.4$ \\
                 & Tag Exponent $\gamma$ & $2.0$ \\
                 & Budget Penalty Decay $\lambda$ & $0.001$ \\
                 & Marginal Penalty $P_{\mathrm{marginal}}$ & $0.25$ \\
                 & EMA Decay $\alpha$ & $0.6$ \\
\bottomrule
\end{tabularx}

\end{table}

\section{RESULTS}

\subsection{Automatic Discovery and Adaptation of Swarm Decomposition}

\begin{figure*}[t]
    \centering
    \vspace{2mm}
    \includegraphics[width=\textwidth]{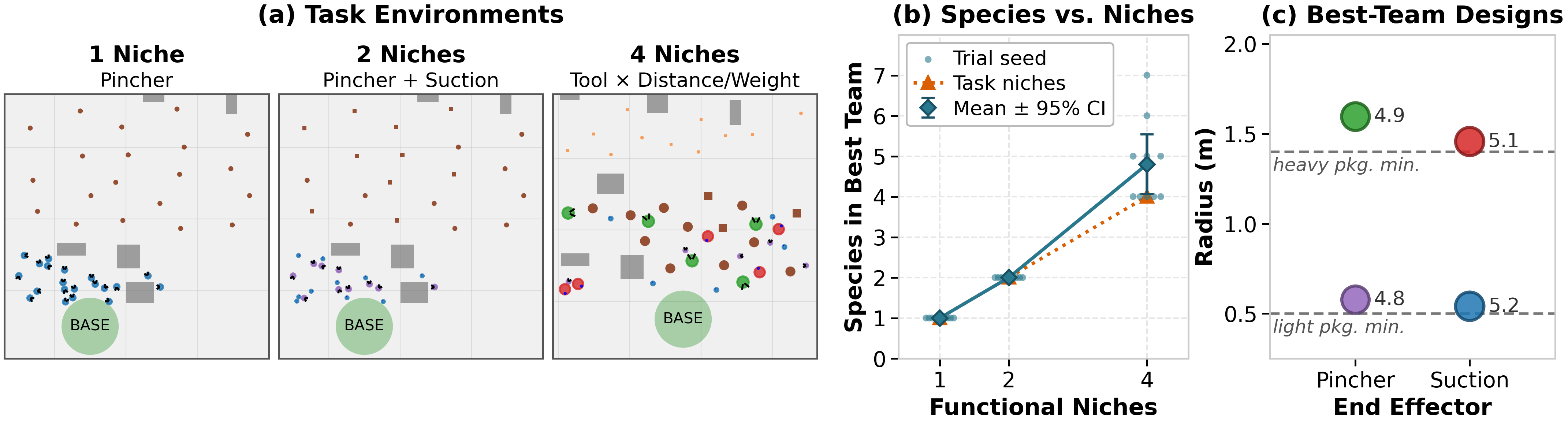}
    \vspace{-6mm}
    \caption{
    Task-dependent emergence of heterogeneous robot designs.
    \textbf{(a)} The three task conditions progressively increase the number of functional niches: a single pincher-only task, a two-tool task requiring pincher and suction robots, and a four-niche task that further separates near/heavy and far/light package roles.
    Each panel shows the best-performing swarm of one representative run.
    Pincher robots have two prongs and suction robots a blue cup.
    Circular and square packages require pincher and suction end effectors, respectively, and darker packages are heavier. 
    Gray rectangles are obstacles.
    \textbf{(b)} Number of species represented in the best-performing deployed swarm across ten independent evolutionary runs for each task condition, with the dotted line indicating the number of functional niches for the task.
    The one- and two-niche tasks consistently produce one and two species, respectively, while the four-niche task produces greater heterogeneity.
    Individual points show results from each run, while markers and error bars show the mean and 95\% confidence interval.
    \textbf{(c)} Radius and end effector of each species in the best-performing swarm of the representative four-niche run shown in (a).
    Colors denote species, matching (a), and marker size and labels give each design's expected number of robots under its evolved relative dominance.
    Dashed lines indicate the minimum radius required to carry light (0.5m) and heavy (1.4m) packages.
    }
    \label{fig:task_complexity}
    \vspace{-5mm}
\end{figure*}

We first evaluate whether the heterogeneous swarm structure discovered by SwarmCoDe changes with the functional requirements of the task.
Figure~\ref{fig:task_complexity}(a) summarizes the one-, two-, and four-niche task conditions, while Fig.~\ref{fig:task_complexity}(b) shows the number of species represented in the best-performing deployed swarm across ten independent evolutionary runs.
The one-niche task consistently produces a single species, while the two-niche task consistently produces two species corresponding to the pincher and suction end-effector requirements.
Increasing the task complexity to four functional niches further increases the evolved heterogeneity, producing an average of 4.8 species in the best-performing swarm.
Because the four-niche experiments impose no cost for maintaining additional robot types, some runs retain multiple species occupying similar functional niches rather than converging to exactly four.
This motivates the fabrication-budget experiments in Section~\ref{sec:budget_results}, where maintaining additional robot designs incurs an explicit cost.

The evolved robot designs also exhibit distinct morphological specialization.
In the representative four-niche solution shown in Fig.~\ref{fig:task_complexity}(c), both pincher and suction robots specialize further into a large design for the heavy packages near the base and a small design for the light packages farther away, with each design evolving a radius close to the minimum required for its package class.
The relative-dominance genes of this solution also evolve a near-even composition of 4.8 to 5.2 expected robots per design.
Across all ten runs, both end-effector types similarly separate into light-package and heavy-package designs.
Thus, the increase in species count corresponds to differences in the physical robot designs and functional roles represented within the deployed swarm.

\begin{figure*}[t]
    \centering
    \vspace{2mm}
    \includegraphics[width=\textwidth]{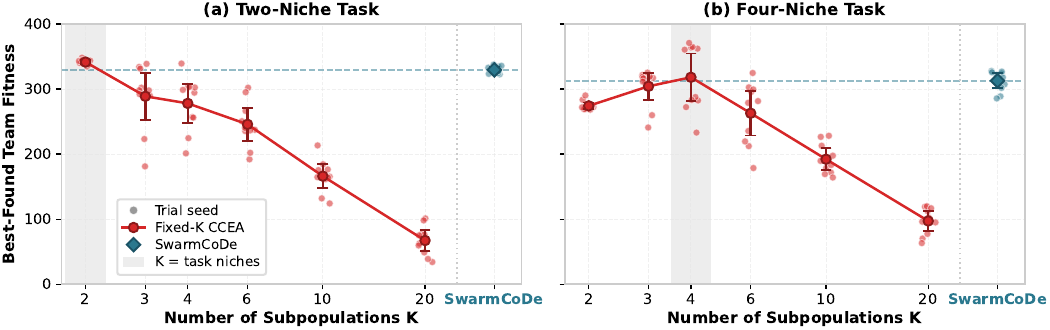}
    \vspace{-5mm}
    \caption{
    Comparison of SwarmCoDe with fixed-$K$ CCEA across the two- and four-niche tasks.
    \textbf{(a)} For the two-niche task, fixed-$K$ CCEA performs best at $K=2$, while performance decreases as the number of predefined subpopulations increases.
    \textbf{(b)} For the four-niche task, fixed-$K$ CCEA performs best at $K=4$, showing that the best choice of $K$ depends on the task.
    Individual points show the best-found team fitness from each of ten independent evolutionary runs, while markers and error bars show the mean and 95\% confidence interval.
    The shaded column marks the value of $K$ equal to the number of task niches, and the dashed line marks the SwarmCoDe mean.
    SwarmCoDe does not require the number of species to be specified before optimization.
    }
    \label{fig:ccea_comparison}
    \vspace{-5mm}
\end{figure*}




These results show that SwarmCoDe can vary the discovered swarm decomposition with the task, but do not establish whether determining this decomposition during optimization provides an advantage over specifying it beforehand.
We therefore compare SwarmCoDe with the fixed-$K$ CCEA baseline described in Section~\ref{sec:implementation_details}.
Figure~\ref{fig:ccea_comparison} shows the best-found team fitness as the number of predefined CCEA subpopulations is varied for both the two- and four-niche tasks.
For the two-niche task, the highest CCEA performance occurs when $K=2$, while increasing $K$ progressively reduces performance.
In contrast, the four-niche task favors a larger decomposition, with the highest mean CCEA performance occurring at $K=4$.
Thus, a fixed-$K$ CCEA can perform very well when an appropriate $K$ is known in advance, but the best choice changes with the task.
SwarmCoDe instead determines the swarm decomposition during optimization and is still able to achieve comparable performance.


The static experiments demonstrate that different task requirements favor different swarm decompositions.
We next test whether this decomposition can adapt when the task itself changes during evolution.
Starting from the two-niche condition, we transition to the four-niche task at generation 250 and continue evolution from the existing population rather than restarting the optimization.


\begin{figure}[t]
    \centering
    \includegraphics[width=\columnwidth]{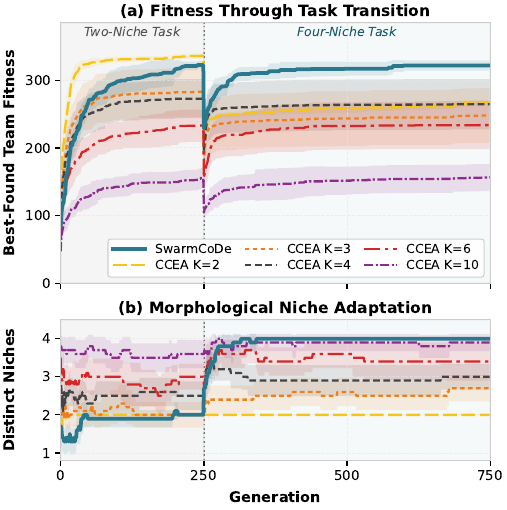}
    \vspace{-5mm}
    \caption{
    Adaptation of SwarmCoDe and fixed-$K$ CCEA following a change in task requirements.
    The task transitions from the two-niche to the four-niche condition at generation 250 (vertical dotted line); the running-best is reset at the transition while evolution continues from the existing population.
    \textbf{(a)} Running-best team fitness.
    \textbf{(b)} Number of distinct morphological niches represented in the running-best team.
    Curves show the mean across ten independent evolutionary runs, with color-matched shading showing the 95\% CI.
    Line colors and styles are shared across both panels.
    }
    \label{fig:dynamic_transition}
    \vspace{-5mm}
\end{figure}

Figure~\ref{fig:dynamic_transition} shows the running-best team fitness for SwarmCoDe and the fixed-\(K\) CCEA baselines before and after the task transition.
During the initial two-niche phase, \(K=2\) achieves the highest fixed-\(K\) CCEA performance, consistent with the static experiments.
Following the transition, the running-best metric is reset for the new task, and each method begins adapting to the four-niche condition.
SwarmCoDe subsequently recovers to the highest-performing solution while simultaneously changing the structure of the deployed swarm.
As shown in Fig.~\ref{fig:dynamic_transition}(b), the best-performing SwarmCoDe teams initially occupy approximately two morphological niches before increasing toward four following the transition.
Thus, rather than requiring the designer to select a new fixed decomposition when the task changes, SwarmCoDe can continue optimization from the existing population while adapting its swarm decomposition to the modified task requirements.

\subsection{Ablation of SwarmCoDe Components}








\begin{figure}[t]
    \centering
    \includegraphics[width=\columnwidth]{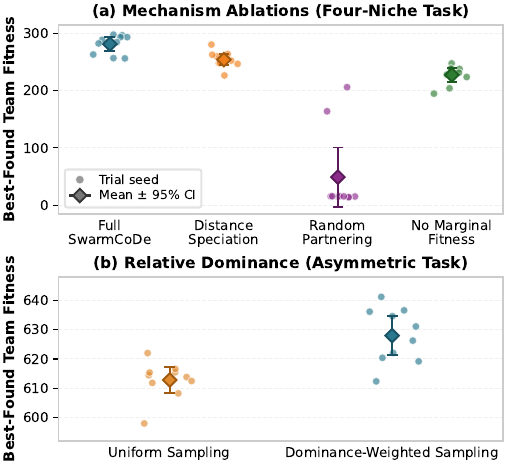}
    \vspace{-5mm}
    \caption{
    Ablations of the main SwarmCoDe mechanisms.
    \textbf{(a)} Peak best-found team fitness across ten independent runs for the full framework and ablations of the speciation metric, partner-selection mechanism, and marginal-contribution fitness on the four-niche task.
    \textbf{(b)} Relative-dominance ablation on an asymmetric two-niche task with 70\% suction and 30\% pincher demand.
    Uniform and dominance-weighted swarm construction are compared in best-found team fitness.
    In both panels, individual points denote evolutionary runs; markers and error bars show the mean and 95\% confidence interval.
    }
    \label{fig:ablations}
    \vspace{-5mm}
\end{figure}








We next evaluate how the main components used by SwarmCoDe for species formation, partner selection, and cooperative credit assignment affect performance.
Figure~\ref{fig:ablations}(a) compares the full framework against three ablations on the four-niche collaborative task.
Distance Speciation replaces the standard species compatibility measure with morphological distance, Random Partnering removes the evolved partner-selection mechanism, and No Marginal Fitness assigns fitness directly from team performance without the focal-versus-baseline marginal evaluation.

The full SwarmCoDe framework achieves the highest mean best-found fitness across the ten independent runs.
Replacing the standard speciation mechanism with distance-based speciation produces a moderate decrease in performance, while removing marginal-contribution fitness results in a larger reduction.
Random Partnering produces the largest decrease, with most runs converging to substantially lower-performing solutions.
These results indicate that the observed performance depends not only on maintaining multiple species, but also on how cooperative species are formed, partnered, and evaluated.

Relative dominance controls a different aspect of the framework by determining the representation of evolved robot designs in the deployed swarm.
To evaluate this mechanism, we use the asymmetric two-niche task, where 70\% of packages require suction and 30\% require pincher robots, while the remaining package properties are held equivalent between the two types.
Figure~\ref{fig:ablations}(b) compares dominance-weighted swarm construction with uniform sampling.
Dominance-weighted sampling evolves an expected composition close to the 70/30 task demand and achieves higher best-found team fitness, while uniform sampling reaches a similar composition only by fielding a second, genetically distinct but worse-performing suction design.
Thus, the relative-dominance gene allows SwarmCoDe to optimize not only which robot designs are represented in the swarm, but also their relative prevalence, without requiring additional designs when the task favors a nonuniform composition.

\subsection{Swarm Co-Design Under Fabrication Constraints}
\label{sec:budget_results}

The previous task-complexity experiments do not impose a direct cost for maintaining additional robot types.
In practice, however, maintaining a larger number of unique robot designs increases fabrication complexity.
We therefore introduce a \$500 fabrication cost for each unique robot design and vary the available swarm budget to evaluate how fabrication constraints affect the evolved swarm design.
These experiments use the four-niche collaborative task, requiring the optimizer to balance the performance benefits of specialization against the additional cost of fabricating multiple robot types.

\begin{figure*}[t]
    \centering
    \vspace{2mm}
    \includegraphics[width=\textwidth]
    {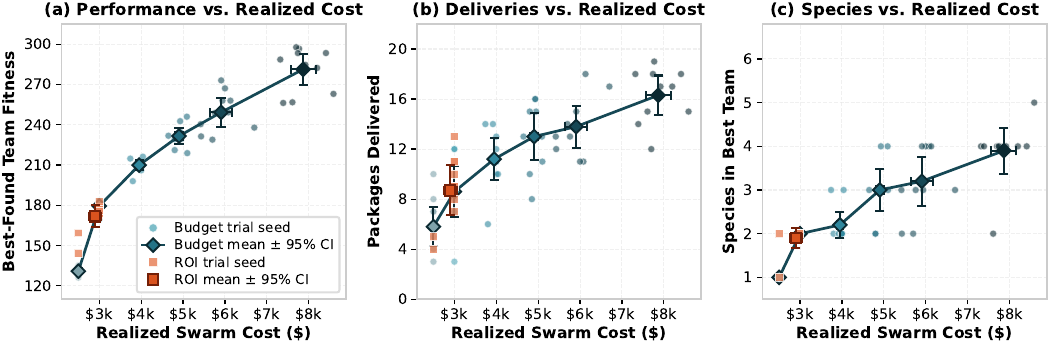}
    \vspace{-2mm}
    \caption{
    Effect of fabrication constraints on the evolved swarm design.
    \textbf{(a)} Best-found team fitness across fabrication-budget conditions, plotted against the realized cost of the resulting swarm.
    \textbf{(b)} Number of packages delivered by the best-performing swarm as fabrication resources increase.
    \textbf{(c)} Number of species represented in the best-performing deployed swarm, showing that larger fabrication budgets allow additional specialized robot designs to be maintained.
    Individual points show results from ten independent evolutionary runs, while markers and error bars show the mean and 95\% confidence interval.
    Darker shades indicate larger fabrication budgets.
    The ROI condition (orange) directly optimizes performance relative to swarm cost without imposing a fixed budget.
    }
    \label{fig:budget_results}
    \vspace{-5mm}
\end{figure*}

Figure~\ref{fig:budget_results} shows the best-performing team across ten independent runs for each budget condition, plotted against the realized cost of the resulting design.
As the available budget increases, SwarmCoDe achieves higher fitness and completes more package deliveries (Fig.~\ref{fig:budget_results}(a-b)).
The number of species represented in the best-performing team also increases with available resources (Fig.~\ref{fig:budget_results}(c)).
Restrictive budgets therefore favor swarms composed of fewer robot types, while additional resources allow more specialized designs to be maintained.
This shows that the degree of swarm heterogeneity is itself part of the co-design tradeoff when additional robot types incur a fabrication cost.

We additionally evaluate a return-on-investment (ROI) objective that directly maximizes performance relative to swarm cost rather than imposing a fixed budget.
The ROI solutions, shown in orange in Fig.~\ref{fig:budget_results}, converge to a lower-cost swarm with approximately two species rather than using the additional resources available in the higher-budget conditions.
The ROI objective therefore favors a lower-cost, less heterogeneous swarm without requiring a fabrication budget to be specified in advance.

\subsection{Scaling Beyond Evolutionary Population Size}
\label{sec:scalability_results}

Finally, we evaluate whether SwarmCoDe can optimize deployed swarms that are substantially larger than the evolutionary population.
We fix the deployed swarm at $N_{\mathrm{swarm}}=200$ and vary the evolutionary population from $N_{\mathrm{pop}}=50$ to 200.

\begin{figure}[t]
    \centering
    \includegraphics[width=\columnwidth]
    {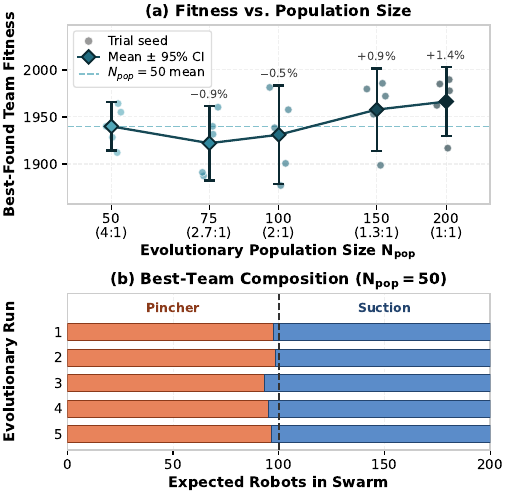}
    \vspace{-5mm}
    \caption{
    Scaling to a 200-robot deployed swarm while varying the evolutionary population size.
    \textbf{(a)} Best-found team fitness for different values of $N_{\mathrm{pop}}$.
    Values in parentheses give the ratio of deployed robots to evolutionary individuals, and percentages show the change in mean fitness relative to $N_{\mathrm{pop}}=50$ (dashed line).
    Individual points show independent evolutionary runs, while markers and error bars show the mean and 95\% confidence interval over the $n=5$ runs.
    \textbf{(b)} Expected composition of the best-found team from each $N_{\mathrm{pop}}=50$ run under its evolved relative-dominance values, colored by end effector.
    The dashed line indicates the task's 50/50 pincher/suction package distribution.
    }
    \label{fig:large_swarm}
    \vspace{-5mm}
\end{figure}

Figure~\ref{fig:large_swarm}(a) shows that reducing the evolutionary population below the deployed swarm size does not produce a corresponding decrease in best-found performance.
Across the tested population sizes, mean fitness remains within approximately 2\% of the $N_{\mathrm{pop}}=50$ condition.
Thus, a high-performing 200-robot swarm does not require an evolutionary population of comparable size: SwarmCoDe can represent the deployed swarm using a smaller set of reusable robot designs, with relative dominance determining their representation in the swarm. 
This decouples deployed swarm size from evolutionary population size, allowing reusable evolved robot designs to represent a substantially larger deployed swarm.


Figure~\ref{fig:large_swarm}(b) shows the expected composition of the best-found team from each of the five $N_{\mathrm{pop}}=50$ runs.
In every run, the best-found team consists of two robot designs, one for each end effector, and the evolved relative-dominance values allocate approximately half of the 200 robots to each design, matching the symmetric package demand.
Each evolved design is therefore instantiated roughly 100 times in the deployed swarm.
Together, these results show that SwarmCoDe can represent and co-design a 200-robot deployed swarm using an evolutionary population substantially smaller than the swarm itself.

\section{CONCLUSIONS}

In this paper, we presented SwarmCoDe, a cooperative co-evolutionary framework that treats the heterogeneous structure of a robot swarm as part of the co-design problem rather than requiring the number of robot species to be specified in advance.
SwarmCoDe uses dynamic speciation, evolved partner selection, and relative dominance to jointly evolve robot designs, task-planning behavior, cooperative relationships, and deployed swarm composition.
Across replicated simulated foraging experiments, SwarmCoDe discovers task-dependent swarm decompositions and adapts them when task requirements change during evolution.
In static comparisons, it achieves performance comparable to the best tested fixed-$K$ CCEA configurations without requiring $K$ to be specified in advance.
We further show that the useful degree of specialization changes with available fabrication resources, with different budgets producing different swarm compositions and robot designs.
Finally, by decoupling deployed swarm composition from evolutionary population size, SwarmCoDe co-designs 200-agent swarms using an evolutionary population of only 50 individuals, demonstrating that the framework can co-design deployed swarms substantially larger than its evolutionary population.
Together, these results demonstrate that heterogeneous swarm structure can itself be treated as an engineering design variable and optimized alongside the robot designs themselves.

Our evaluation remains limited to simulated 2D foraging tasks, and future work should investigate how these findings transfer to physical robot swarms and broader task families. We also plan to extend the framework to more tightly coupled design domains, such as low-level dynamics and control that directly depend on morphology, as well as the co-design of sensing capabilities and decentralized communication. Finally, we aim to investigate algorithmic techniques for reducing evaluation cost and per-generation fitness variance.





\section*{ACKNOWLEDGMENT}

The authors utilized generative AI tools (Gemini, Claude, and ChatGPT) for code generation and manuscript editing. 
All AI-assisted code and manuscript content were reviewed and verified by the authors.


\balance

\bibliographystyle{IEEEtran}
\bibliography{control, references}

@IEEEtranBSTCTL{IEEEexample:BSTcontrol,
  CTLuse_url = "no",
  CTLuse_forced_etal = "no"
}

@article{cheneyScalableCooptimizationMorphology2018,
	title = {Scalable co-optimization of morphology and control in embodied machines},
	volume = {15},
	issn = {1742-5689, 1742-5662},
	url = {https://royalsocietypublishing.org/doi/10.1098/rsif.2017.0937},
	doi = {10.1098/rsif.2017.0937},
	language = {en},
	number = {143},
	urldate = {2026-10-21},
	journal = {Journal of The Royal Society Interface},
	author = {Cheney, Nick and Bongard, Josh and SunSpiral, Vytas and Lipson, Hod},
	month = jun,
	year = {2018},
	pages = {20170937},
}

@article{francescaAutoMoDeNovelApproach2014,
	title = {{AutoMoDe}: {A} novel approach to the automatic design of control software for robot swarms},
	volume = {8},
	copyright = {http://www.springer.com/tdm},
	issn = {1935-3812, 1935-3820},
	shorttitle = {{AutoMoDe}},
	url = {http://link.springer.com/10.1007/s11721-014-0092-4},
	doi = {10.1007/s11721-014-0092-4},
	language = {en},
	number = {2},
	urldate = {2026-10-21},
	journal = {Swarm Intelligence},
	author = {Francesca, Gianpiero and Brambilla, Manuele and Brutschy, Arne and Trianni, Vito and Birattari, Mauro},
	month = jun,
	year = {2014},
	pages = {89--112},
}

@article{salmanConcurrentDesignControl2019,
	title = {Concurrent design of control software and configuration of hardware for robot swarms under economic constraints},
	volume = {5},
	copyright = {https://creativecommons.org/licenses/by/4.0/},
	issn = {2376-5992},
	url = {https://peerj.com/articles/cs-221},
	doi = {10.7717/peerj-cs.221},
	language = {en},
	urldate = {2026-10-21},
	journal = {PeerJ Computer Science},
	author = {Salman, Muhammad and Ligot, Antoine and Birattari, Mauro},
	month = sep,
	year = {2019},
	pages = {e221},
}

@article{pughQualityDiversityNew2016,
	title = {Quality {Diversity}: {A} {New} {Frontier} for {Evolutionary} {Computation}},
	volume = {3},
	issn = {2296-9144},
	shorttitle = {Quality {Diversity}},
	url = {http://journal.frontiersin.org/Article/10.3389/frobt.2016.00040/abstract},
	doi = {10.3389/frobt.2016.00040},
	urldate = {2026-10-21},
	journal = {Frontiers in Robotics and AI},
	author = {Pugh, Justin K. and Soros, Lisa B. and Stanley, Kenneth O.},
	month = jul,
	year = {2016},
}

@article{omidvarCooperativeCoEvolutionDifferential2014,
	title = {Cooperative {Co}-{Evolution} {With} {Differential} {Grouping} for {Large} {Scale} {Optimization}},
	volume = {18},
	copyright = {https://ieeexplore.ieee.org/Xplorehelp/downloads/license-information/IEEE.html},
	issn = {1089-778X, 1089-778X, 1941-0026},
	url = {https://ieeexplore.ieee.org/document/6595612/},
	doi = {10.1109/TEVC.2013.2281543},
	number = {3},
	urldate = {2026-10-21},
	journal = {IEEE Transactions on Evolutionary Computation},
	author = {Omidvar, Mohammad Nabi and Li, Xiaodong and Mei, Yi and Yao, Xin},
	month = jun,
	year = {2014},
	pages = {378--393},
}

@article{waibelGeneticTeamComposition2009,
	title = {Genetic {Team} {Composition} and {Level} of {Selection} in the {Evolution} of {Cooperation}},
	volume = {13},
	copyright = {https://ieeexplore.ieee.org/Xplorehelp/downloads/license-information/IEEE.html},
	issn = {1089-778X},
	url = {http://ieeexplore.ieee.org/document/5089892/},
	doi = {10.1109/TEVC.2008.2011741},
	number = {3},
	urldate = {2026-10-21},
	journal = {IEEE Transactions on Evolutionary Computation},
	author = {Waibel, Markus and Keller, Laurent and Floreano, Dario},
	month = jun,
	year = {2009},
	pages = {648--660},
}

@article{dorigoSwarmanoidNovelConcept2013,
	title = {Swarmanoid: {A} {Novel} {Concept} for the {Study} of {Heterogeneous} {Robotic} {Swarms}},
	volume = {20},
	copyright = {https://ieeexplore.ieee.org/Xplorehelp/downloads/license-information/IEEE.html},
	issn = {1070-9932},
	shorttitle = {Swarmanoid},
	url = {http://ieeexplore.ieee.org/document/6603259/},
	doi = {10.1109/MRA.2013.2252996},
	number = {4},
	urldate = {2026-10-21},
	journal = {IEEE Robotics \& Automation Magazine},
	author = {Dorigo, Marco and Floreano, Dario and Gambardella, Luca Maria and Mondada, Francesco and Nolfi, Stefano and Baaboura, Tarek and Birattari, Mauro and Bonani, Michael and Brambilla, Manuele and Brutschy, Arne and Burnier, Daniel and Campo, Alexandre and Christensen, Anders Lyhne and Decugniere, Antal and Di Caro, Gianni and Ducatelle, Frederick and Ferrante, Eliseo and Forster, Alexander and Gonzales, Javier Martinez and Guzzi, Jerome and Longchamp, Valentin and Magnenat, Stephane and Mathews, Nithin and Montes De Oca, Marco and O'Grady, Rehan and Pinciroli, Carlo and Pini, Giovanni and Retornaz, Philippe and Roberts, James and Sperati, Valerio and Stirling, Timothy and Stranieri, Alessandro and Stutzle, Thomas and Trianni, Vito and Tuci, Elio and Turgut, Ali Emre and Vaussard, Florian},
	month = dec,
	year = {2013},
	pages = {60--71},
}

@article{kucklingRecentTrendsRobot2023,
	title = {Recent trends in robot learning and evolution for swarm robotics},
	volume = {10},
	issn = {2296-9144},
	url = {https://www.frontiersin.org/articles/10.3389/frobt.2023.1134841/full},
	doi = {10.3389/frobt.2023.1134841},
	urldate = {2026-10-21},
	journal = {Frontiers in Robotics and AI},
	author = {Kuckling, Jonas},
	month = apr,
	year = {2023},
	pages = {1134841},
}

@article{birattariAutomaticOffLineDesign2019,
	title = {Automatic {Off}-{Line} {Design} of {Robot} {Swarms}: {A} {Manifesto}},
	volume = {6},
	issn = {2296-9144},
	shorttitle = {Automatic {Off}-{Line} {Design} of {Robot} {Swarms}},
	url = {https://www.frontiersin.org/article/10.3389/frobt.2019.00059/full},
	doi = {10.3389/frobt.2019.00059},
	urldate = {2026-10-21},
	journal = {Frontiers in Robotics and AI},
	author = {Birattari, Mauro and Ligot, Antoine and Bozhinoski, Darko and Brambilla, Manuele and Francesca, Gianpiero and Garattoni, Lorenzo and Garzón Ramos, David and Hasselmann, Ken and Kegeleirs, Miquel and Kuckling, Jonas and Pagnozzi, Federico and Roli, Andrea and Salman, Muhammad and Stützle, Thomas},
	month = jul,
	year = {2019},
	pages = {59},
}

@incollection{vandiggelenEmergenceSpecialisedCollective2024,
	address = {Cham},
	title = {Emergence of {Specialised} {Collective} {Behaviors} in {Evolving} {Heterogeneous} {Swarms}},
	volume = {15149},
	isbn = {978-3-031-70067-5 978-3-031-70068-2},
	url = {https://link.springer.com/10.1007/978-3-031-70068-2_4},
	doi = {10.1007/978-3-031-70068-2_4},
	language = {en},
	urldate = {2026-01-26},
	booktitle = {Parallel {Problem} {Solving} from {Nature} - {PPSN} {XVIII}},
	publisher = {Springer Nature Switzerland},
	author = {Van Diggelen, Fuda and De Carlo, Matteo and Cambier, Nicolas and Ferrante, Eliseo and Eiben, Guszti},
	editor = {Affenzeller, Michael and Winkler, Stephan M. and Kononova, Anna V. and Trautmann, Heike and Tušar, Tea and Machado, Penousal and Bäck, Thomas},
	year = {2024},
	pages = {53--69},
}

@incollection{eibenCollectiveSpecializationEvolutionary2007,
	address = {Berlin, Heidelberg},
	title = {Collective {Specialization} for {Evolutionary} {Design} of a {Multi}-robot {System}},
	volume = {4433},
	isbn = {978-3-540-71540-5 978-3-540-71541-2},
	url = {http://link.springer.com/10.1007/978-3-540-71541-2_13},
	doi = {10.1007/978-3-540-71541-2_13},
	language = {en},
	urldate = {2026-01-22},
	booktitle = {Swarm {Robotics}},
	publisher = {Springer Berlin Heidelberg},
	author = {Eiben, Agoston E. and Nitschke, Geoff S. and Schut, Martijn C.},
	editor = {Sahin, Erol and Spears, William M. and Winfield, Alan F. T.},
	year = {2007},
	pages = {189--205},
}

@article{haComputationalDesignRobotic2018,
	title = {Computational {Design} of {Robotic} {Devices} {From} {High}-{Level} {Motion} {Specifications}},
	volume = {34},
	issn = {1941-0468},
	url = {https://ieeexplore.ieee.org/document/8395009/?arnumber=8395009},
	doi = {10.1109/TRO.2018.2830419},
	number = {5},
	urldate = {2025-03-20},
	journal = {IEEE Transactions on Robotics},
	author = {Ha, Sehoon and Coros, Stelian and Alspach, Alexander and Bern, James M. and Kim, Joohyung and Yamane, Katsu},
	month = oct,
	year = {2018},
	pages = {1240--1251},
}

@article{agoginoEfficientEvaluationFunctions2008,
	title = {Efficient evaluation functions for evolving coordination},
	volume = {16},
	number = {2},
	journal = {Evolutionary Computation},
	publisher = {MIT Press One Rogers Street, Cambridge, MA 02142-1209, USA journals-info …},
	author = {Agogino, Adrian and Tumer, Kagan},
	year = {2008},
	pages = {257--288},
}

@article{wolpertOptimalPayoffFunctions2001,
	title = {Optimal payoff functions for members of collectives},
	volume = {4},
	number = {02n03},
	journal = {Advances in Complex Systems},
	publisher = {World Scientific},
	author = {Wolpert, David H and Tumer, Kagan},
	year = {2001},
	pages = {265--279},
}

@article{yongCoevolutionRoleBasedCooperation2009,
	title = {Coevolution of {Role}-{Based} {Cooperation} in {Multiagent} {Systems}},
	volume = {1},
	issn = {1943-0612},
	url = {https://ieeexplore.ieee.org/document/5345731/},
	doi = {10.1109/TAMD.2009.2037732},
	number = {3},
	urldate = {2026-10-09},
	journal = {IEEE Transactions on Autonomous Mental Development},
	author = {Yong, Chern Han and Miikkulainen, Risto},
	month = oct,
	year = {2009},
	pages = {170--186},
}

@article{potterCooperativeCoevolutionArchitecture2000,
	title = {Cooperative {Coevolution}: {An} {Architecture} for {Evolving} {Coadapted} {Subcomponents}},
	volume = {8},
	issn = {1063-6560, 1530-9304},
	shorttitle = {Cooperative {Coevolution}},
	url = {https://direct.mit.edu/evco/article/8/1/1-29/859},
	doi = {10.1162/106365600568086},
	language = {en},
	number = {1},
	urldate = {2026-10-09},
	journal = {Evolutionary Computation},
	author = {Potter, Mitchell A. and Jong, Kenneth A. De},
	month = mar,
	year = {2000},
	pages = {1--29},
}

@inproceedings{potterEvolvingNeuralNetworks1995,
	title = {Evolving neural networks with collaborative species},
	booktitle = {Summer {Computer} {Simulation} {Conference}},
	publisher = {SOCIETY FOR COMPUTER SIMULATION, ETC},
	author = {Potter, Mitchell A and De Jong, Kenneth A},
	year = {1995},
	pages = {340--345},
}

@article{rioloEvolutionCooperationReciprocity2001,
	title = {Evolution of cooperation without reciprocity},
	volume = {414},
	copyright = {http://www.springer.com/tdm},
	issn = {0028-0836, 1476-4687},
	url = {https://www.nature.com/articles/35106555},
	doi = {10.1038/35106555},
	language = {en},
	number = {6862},
	urldate = {2026-10-09},
	journal = {Nature},
	author = {Riolo, Rick L. and Cohen, Michael D. and Axelrod, Robert},
	month = nov,
	year = {2001},
	pages = {441--443},
}

@article{nitschkeEvolvingTeamBehaviors2012,
	title = {Evolving team behaviors with specialization},
	volume = {13},
	copyright = {http://www.springer.com/tdm},
	issn = {1389-2576, 1573-7632},
	url = {http://link.springer.com/10.1007/s10710-012-9166-5},
	doi = {10.1007/s10710-012-9166-5},
	language = {en},
	number = {4},
	urldate = {2026-07-17},
	journal = {Genetic Programming and Evolvable Machines},
	author = {Nitschke, G. S. and Eiben, A. E. and Schut, M. C.},
	month = dec,
	year = {2012},
	pages = {493--536},
}

@inproceedings{wilhelmMonotoneSubsystemDecomposition2025,
	title = {Monotone {Subsystem} {Decomposition} for {Efficient} {Multi}-{Objective} {Robot} {Design}},
	url = {https://ieeexplore.ieee.org/document/11128384/},
	doi = {10.1109/ICRA55743.2025.11128384},
	urldate = {2026-05-07},
	booktitle = {2025 {IEEE} {International} {Conference} on {Robotics} and {Automation} ({ICRA})},
	author = {Wilhelm, Andrew and Napp, Nils},
	month = may,
	year = {2025},
	pages = {8114--8120},
}

@article{brambillaSwarmRoboticsReview2013,
	title = {Swarm robotics: a review from the swarm engineering perspective},
	volume = {7},
	issn = {1935-3820},
	shorttitle = {Swarm robotics},
	url = {https://doi.org/10.1007/s11721-012-0075-2},
	doi = {10.1007/s11721-012-0075-2},
	language = {en},
	number = {1},
	urldate = {2026-03-05},
	journal = {Swarm Intelligence},
	author = {Brambilla, Manuele and Ferrante, Eliseo and Birattari, Mauro and Dorigo, Marco},
	month = mar,
	year = {2013},
	pages = {1--41},
}

@inproceedings{dambrosioGenerativeEncodingMultiagent2008,
	address = {New York, NY, USA},
	series = {{GECCO} '08},
	title = {Generative encoding for multiagent learning},
	isbn = {978-1-60558-130-9},
	url = {https://dl.acm.org/doi/10.1145/1389095.1389256},
	doi = {10.1145/1389095.1389256},
	urldate = {2026-02-05},
	booktitle = {Proceedings of the 10th annual conference on {Genetic} and evolutionary computation},
	publisher = {Association for Computing Machinery},
	author = {D'Ambrosio, David B. and Stanley, Kenneth O.},
	month = jul,
	year = {2008},
	pages = {819--826},
}

@article{magnussenMulticopterDesignOptimization2015,
	title = {Multicopter {Design} {Optimization} and {Validation}},
	volume = {36},
	issn = {0332-7353, 1890-1328},
	url = {http://www.mic-journal.no/ABS/MIC-2015-2-1.asp},
	doi = {10.4173/mic.2015.2.1},
	number = {2},
	urldate = {2024-08-14},
	journal = {Modeling, Identification and Control: A Norwegian Research Bulletin},
	author = {Magnussen, Oyvind and Ottestad, Morten and Hovland, Geir},
	year = {2015},
	pages = {67--79},
}

@inproceedings{zardiniCoDesignAVEnabledMobility2020,
	title = {On the {Co}-{Design} of {AV}-{Enabled} {Mobility} {Systems}},
	url = {http://arxiv.org/abs/2003.04739},
	doi = {10.1109/ITSC45102.2020.9294499},
	urldate = {2026-03-02},
	booktitle = {2020 {IEEE} 23rd {International} {Conference} on {Intelligent} {Transportation} {Systems} ({ITSC})},
	author = {Zardini, Gioele and Lanzetti, Nicolas and Salazar, Mauro and Censi, Andrea and Frazzoli, Emilio and Pavone, Marco},
	month = sep,
	year = {2020},
	note = {arXiv:2003.04739 [eess]},
	pages = {1--8},
}

@article{ferranteEvolutionSelfOrganizedTask2015,
	title = {Evolution of {Self}-{Organized} {Task} {Specialization} in {Robot} {Swarms}},
	volume = {11},
	issn = {1553-7358},
	url = {https://dx.plos.org/10.1371/journal.pcbi.1004273},
	doi = {10.1371/journal.pcbi.1004273},
	language = {en},
	number = {8},
	urldate = {2026-02-27},
	journal = {PLOS Computational Biology},
	author = {Ferrante, Eliseo and Turgut, Ali Emre and Duéñez-Guzmán, Edgar and Dorigo, Marco and Wenseleers, Tom},
	editor = {Sporns, Olaf},
	month = aug,
	year = {2015},
	pages = {e1004273},
}

@article{gomesDynamicTeamHeterogeneity2018,
	title = {Dynamic {Team} {Heterogeneity} in {Cooperative} {Coevolutionary} {Algorithms}},
	volume = {22},
	issn = {1941-0026},
	url = {https://ieeexplore.ieee.org/document/8141987/},
	doi = {10.1109/TEVC.2017.2779840},
	number = {6},
	urldate = {2026-02-05},
	journal = {IEEE Transactions on Evolutionary Computation},
	author = {Gomes, Jorge and Mariano, Pedro and Christensen, Anders Lyhne},
	month = dec,
	year = {2018},
	pages = {934--948},
}

@article{gomesEvolutionSwarmRobotics2013,
	title = {Evolution of swarm robotics systems with novelty search},
	volume = {7},
	copyright = {http://www.springer.com/tdm},
	issn = {1935-3812, 1935-3820},
	url = {http://link.springer.com/10.1007/s11721-013-0081-z},
	doi = {10.1007/s11721-013-0081-z},
	language = {en},
	number = {2-3},
	urldate = {2026-02-05},
	journal = {Swarm Intelligence},
	author = {Gomes, Jorge and Urbano, Paulo and Christensen, Anders Lyhne},
	month = sep,
	year = {2013},
	pages = {115--144},
}

@article{liCoevolutionFrameworkSwarm2015,
	title = {Co-evolution framework of swarm self-assembly robots},
	volume = {148},
	issn = {0925-2312},
	url = {https://www.sciencedirect.com/science/article/pii/S0925231214009394},
	doi = {10.1016/j.neucom.2012.10.047},
	urldate = {2026-01-27},
	journal = {Neurocomputing},
	author = {Li, Haiyuan and Wei, Hongxing and Xiao, Jiangyang and Wang, Tianmiao},
	month = jan,
	year = {2015},
	pages = {112--121},
}

@article{maSurveyCooperativeCoEvolutionary2019,
	title = {A {Survey} on {Cooperative} {Co}-{Evolutionary} {Algorithms}},
	volume = {23},
	copyright = {https://ieeexplore.ieee.org/Xplorehelp/downloads/license-information/IEEE.html},
	issn = {1089-778X, 1089-778X, 1941-0026},
	url = {https://ieeexplore.ieee.org/document/8454482/},
	doi = {10.1109/TEVC.2018.2868770},
	language = {en},
	number = {3},
	urldate = {2026-01-26},
	journal = {IEEE Transactions on Evolutionary Computation},
	author = {Ma, Xiaoliang and Li, Xiaodong and Zhang, Qingfu and Tang, Ke and Liang, Zhengping and Xie, Weixin and Zhu, Zexuan},
	month = jun,
	year = {2019},
	pages = {421--441},
}

@inproceedings{liSpeciesBasedEvolutionary2010,
	title = {Species based evolutionary algorithms for multimodal optimization: {A} brief review},
	shorttitle = {Species based evolutionary algorithms for multimodal optimization},
	url = {https://ieeexplore.ieee.org/document/5586349/},
	doi = {10.1109/CEC.2010.5586349},
	urldate = {2026-01-22},
	booktitle = {{IEEE} {Congress} on {Evolutionary} {Computation}},
	author = {Li, Jian-Ping and Li, Xiao-Dong and Wood, Alastair},
	month = jul,
	year = {2010},
	pages = {1--8},
}

@article{spielbergLearningInTheLoopOptimizationEndToEnd2019,
	title = {Learning-{In}-{The}-{Loop} {Optimization}: {End}-{To}-{End} {Control} {And} {Co}-{Design} {Of} {Soft} {Robots} {Through} {Learned} {Deep} {Latent} {Representations}},
	language = {en},
	journal = {Advances in Neural Information Processing Systems},
	author = {Spielberg, Andrew and Zhao, Allan and Hu, Yuanming and Du, Tao and Matusik, Wojciech and Rus, Daniela},
	year = {2019},
}

@inproceedings{spielbergFunctionalCooptimizationArticulated2017,
	address = {Singapore, Singapore},
	title = {Functional co-optimization of articulated robots},
	isbn = {978-1-5090-4633-1},
	url = {http://ieeexplore.ieee.org/document/7989587/},
	doi = {10.1109/ICRA.2017.7989587},
	language = {en},
	urldate = {2025-09-17},
	booktitle = {2017 {IEEE} {International} {Conference} on {Robotics} and {Automation} ({ICRA})},
	publisher = {IEEE},
	author = {Spielberg, Andrew and Araki, Brandon and Sung, Cynthia and Tedrake, Russ and Rus, Daniela},
	month = may,
	year = {2017},
	pages = {5035--5042},
}

@article{stanleyEvolvingNeuralNetworks2002,
	title = {Evolving {Neural} {Networks} through {Augmenting} {Topologies}},
	volume = {10},
	issn = {1063-6560, 1530-9304},
	url = {https://direct.mit.edu/evco/article/10/2/99-127/1123},
	doi = {10.1162/106365602320169811},
	language = {en},
	number = {2},
	urldate = {2025-09-17},
	journal = {Evolutionary Computation},
	author = {Stanley, Kenneth O. and Miikkulainen, Risto},
	month = jun,
	year = {2002},
	pages = {99--127},
}

@article{milojevicCODEIResourceEfficientTaskDriven2025,
	title = {{CODEI}: {Resource}-{Efficient} {Task}-{Driven} {Co}-{Design} of {Perception} and {Decision} {Making} for {Mobile} {Robots} {Applied} to {Autonomous} {Vehicles}},
	copyright = {https://ieeexplore.ieee.org/Xplorehelp/downloads/license-information/IEEE.html},
	issn = {1552-3098, 1941-0468},
	shorttitle = {{CODEI}},
	url = {https://ieeexplore.ieee.org/document/10930687/},
	doi = {10.1109/TRO.2025.3552347},
	language = {en},
	urldate = {2025-03-20},
	journal = {IEEE Transactions on Robotics},
	author = {Milojevic, Dejan and Zardini, Gioele and Elser, Miriam and Censi, Andrea and Frazzoli, Emilio},
	year = {2025},
	pages = {1--20},
}

@article{lipsonAutomaticDesignManufacture2000,
	title = {Automatic design and manufacture of robotic lifeforms},
	volume = {406},
	copyright = {2000 Macmillan Magazines Ltd.},
	issn = {1476-4687},
	url = {https://www.nature.com/articles/35023115},
	doi = {10.1038/35023115},
	language = {en},
	number = {6799},
	urldate = {2024-11-15},
	journal = {Nature},
	publisher = {Nature Publishing Group},
	author = {Lipson, Hod and Pollack, Jordan B.},
	month = aug,
	year = {2000},
	pages = {974--978},
}

@inproceedings{wilhelmConstraintProgrammingComponentLevel2023,
	title = {Constraint {Programming} for {Component}-{Level} {Robot} {Design}},
	issn = {2153-0866},
	url = {https://ieeexplore.ieee.org/document/10341679/?arnumber=10341679},
	doi = {10.1109/IROS55552.2023.10341679},
	urldate = {2024-09-11},
	booktitle = {2023 {IEEE}/{RSJ} {International} {Conference} on {Intelligent} {Robots} and {Systems} ({IROS})},
	author = {Wilhelm, Andrew and Napp, Nils},
	month = oct,
	year = {2023},
	pages = {460--466},
}

@inproceedings{johanssonComponentBasedOptimization2007,
	address = {Las Vegas, Nevada, USA},
	title = {A {Component} {Based} {Optimization} {Approach} for {Modular} {Robot} {Design}},
	isbn = {978-0-7918-4807-4},
	url = {https://asmedigitalcollection.asme.org/IDETC-CIE/proceedings/IDETC-CIE2007/48078/911/330313},
	doi = {10.1115/DETC2007-35329},
	language = {en},
	urldate = {2024-09-11},
	booktitle = {Volume 6: 33rd {Design} {Automation} {Conference}, {Parts} {A} and {B}},
	publisher = {ASMEDC},
	author = {Johansson, Bjo¨rn and Pettersson, Marcus and O¨lvander, Johan},
	month = jan,
	year = {2007},
	pages = {911--920},
}

@inproceedings{carloneRobotCodesignMonotone2019,
	title = {Robot {Co}-design: {Beyond} the {Monotone} {Case}},
	issn = {2577-087X},
	shorttitle = {Robot {Co}-design},
	url = {https://ieeexplore.ieee.org/document/8793926/?arnumber=8793926},
	doi = {10.1109/ICRA.2019.8793926},
	urldate = {2024-08-14},
	booktitle = {2019 {International} {Conference} on {Robotics} and {Automation} ({ICRA})},
	author = {Carlone, Luca and Pinciroli, Carlo},
	month = may,
	year = {2019},
	pages = {3024--3030},
}

@article{zhaoRoboGrammarGraphGrammar2020,
	title = {{RoboGrammar}: {Graph} grammar for terrain-optimized robot design},
	volume = {39},
	issn = {15577368},
	doi = {10.1145/3414685.3417831},
	number = {6},
	journal = {ACM Transactions on Graphics},
	publisher = {Association for Computing Machinery},
	author = {Zhao, Allan and Xu, Jie and Konaković-Luković, Mina and Hughes, Josephine and Spielberg, Andrew and Rus, Daniela and Matusik, Wojciech},
	month = nov,
	year = {2020},
}

\end{document}